\newif\ifarxiv
\arxivtrue  %

  \documentclass[accepted]{uai2023} %

\usepackage[american]{babel}

\usepackage{natbib} %
\usepackage{mathtools} %
\usepackage{booktabs} %
\usepackage{tikz} %

\usepackage[capitalise]{cleveref}
\usepackage{caption}
\usepackage{subcaption}
\usepackage{pgfplots}

\usepackage{pifont}%

\usepackage{xargs}
\usepackage[colorinlistoftodos,textsize=normalsize]{todonotes} %
\newcommandx{\todoc}[2][1=]{{\todo[linecolor=orange,backgroundcolor=orange!25,bordercolor=orange,#1]{
      TODO: #2}}}
\newcommandx{\unsure}[2][1=]{{\todo[linecolor=yellow,backgroundcolor=yellow!25,bordercolor=yellow,#1]{
      UNSURE: #2}}}
\newcommandx{\change}[2][1=]{{\todo[linecolor=blue,backgroundcolor=blue!25,bordercolor=blue,#1]{
      CHANGE: #2}}}
\newcommandx{\info}[2][1=]{{\todo[linecolor=green,backgroundcolor=green!25,bordercolor=green,#1]{
      INFO: #2}}}
\newcommandx{\improvement}[2][1=]{{\todo[linecolor=violet,backgroundcolor=violet!25,bordercolor=violet,#1]{
      IMPROVEMENT: #2}}}
\newcommandx{\thiswillnotshow}[2][1=]{{\todo[disable,#1]{THIS WILL NOT SHOW:
      #2}}}

\usepackage{bm}

\usepackage{algorithm} \usepackage[noend]{algpseudocode}

\usepackage{amsmath,amssymb}
\DeclareMathOperator*{\argmax}{arg\,max}

\usetikzlibrary{colorbrewer} \usetikzlibrary{positioning}
\usetikzlibrary{shapes.geometric} \usetikzlibrary{arrows}
\usetikzlibrary{arrows.meta} \usetikzlibrary{patterns,decorations.pathreplacing}
\usetikzlibrary{fit} \usetikzlibrary{backgrounds}
\usetikzlibrary{shapes,backgrounds,calc} \usetikzlibrary{shadows}
\usetikzlibrary{patterns} \definecolor{mygray}{RGB}{190,190,190}

\newcommand{\node}{\mathsf{N}}
\newcommand{\nodeD}{\node}
\newcommand{\ndi}{\nodeD_i}
\newcommand{\ndj}{\nodeD_j}
\newcommand{\sumnode}{\mathsf{S}}
\newcommand{\si}{\sumnode_i}
\newcommand{\sj}{\sumnode_j}
\newcommand{\sk}{\sumnode_k}
\newcommand{\prodnode}{\mathsf{P}}
\newcommand{\leafnode}{\mathsf{L}}
\newcommand{\sumnodeD}{\sumnode}  %
\newcommand{\prodnodeD}{\prodnode}  %
\newcommand{\leafnodeD}{\leafnode}  %

\newcommand{\E}[1]{\mathbb{E}\!\left[#1\right]}
\newcommand{\Esi}{\E{\sumnode_i}}
\newcommand{\Esj}{\E{\sumnode_j}}

\newcommand{\Ens}[1]{\mathbb{E}\left[#1\right]}  %
\newcommand{\Var}[1]{\text{Var}\!\left[#1\right]}
\newcommand{\Varns}[1]{\text{Var}\left[#1\right]} %
\newcommand{\Cov}[1]{\text{Cov}\!\left[#1\right]}
\newcommand{\Covns}[1]{\text{Cov}\left[#1\right]} %

\newcommand{\Einline}[1]{\mathbb{E}[#1]}
\newcommand{\Varinline}[1]{\text{Var}[#1]}
\newcommand{\Covinline}[1]{\text{Cov}[#1]}

\newcommand{\fn}[2]{#1\! \left( #2 \right)}

\newcommand{\ci}{c_i}  %
\newcommand{\cj}{c_j}  %

\newcommand{\cbar}{\,|\,}
\newcommand{\X}{\mathbf{X}}

\newcommand{\Z}{\mathbf{Z}}
\newcommand{\x}{\mathbf{x}}

\newcommand{\data}{\mathcal{D}}

\newcommand{\ind}{\perp\!\!\!\!\perp} %

\newcommand{\spn}{\mathcal{S}}

\newcommand{\scope}[1]{\mathbf{sc}\left(#1\right)}

\newcommand{\ch}{\mathbf{ch}}

\newcommand{\mathbold}[1]{\boldsymbol{#1}}

\newcommand{\mbx}{\mathbold{x}}

\newcommand{\mbtheta}{\mathbold{\theta}}

\newlength{\minnodesize}
\newlength{\nodethickness}
\newlength{\nodedist}
\newlength{\prodnodedist}
\newlength{\leafnodedist}
\definecolor{tab10green}{HTML}{2CA02C}
\definecolor{tab10blue}{HTML}{1f77b4}
\definecolor{tab10red}{HTML}{d62728}

\title{Probabilistic Circuits That Know What They Don't Know}

\author[1]{Fabrizio Ventola*}
\author[1]{Steven Braun*}
\author[1]{Zhongjie Yu}
\author[1,2]{Martin Mundt}
\author[1,2,3,4]{Kristian Kersting}
\affil[1]{%
    Department of Computer Science\\
    TU Darmstadt\\
    Darmstadt, Germany
}
\affil[2]{%
    Hessian Center for AI (hessian.AI)\\
    Darmstadt, Germany
}
\affil[3]{%
    German Research Center for Artificial Intelligence (DFKI)\\
    Darmstadt, Germany
}
\affil[4]{%
    Centre for Cognitive Science\\
    TU Darmstadt\\
    Darmstadt, Germany
}

\makeatletter
\def\blfootnote{\gdef\@thefnmark{}\@footnotetext}
\makeatother

\begin{document}
\maketitle

\begin{abstract}
Probabilistic circuits (PCs) are models that allow exact and tractable
probabilistic inference. In contrast to neural networks, they are often assumed
to be well-calibrated and robust to out-of-distribution (OOD) data. In this
paper, we show that PCs are in fact not robust to OOD data, i.e., they don't
know what they don't know. 
We then show how this challenge can be overcome by model uncertainty quantification. To this end, we
propose tractable dropout inference (TDI), an inference procedure to estimate
uncertainty by deriving an analytical solution to Monte Carlo dropout (MCD)
through variance propagation. Unlike MCD in neural networks,
which comes at the cost of multiple network evaluations, TDI provides
tractable sampling-free uncertainty estimates in a single forward pass.
TDI improves the robustness of PCs to distribution shift and OOD data,
demonstrated through a series of experiments evaluating the classification
confidence and uncertainty estimates on real-world data.
\end{abstract}

\section{Introduction}
\blfootnote{*indicates equal contribution}
The majority of modern machine learning research concentrates on a closed-world setting \citep{boult_2019}.
Here, the value of a model is judged by its performance on a dedicated train-validation-test split from a joint data distribution. Such a focus discounts crucial requirements for real-world inference, where data with a shift in distribution, conceptually novel data, or various combinations of unfiltered corruptions and perturbations are typically encountered \citep{boult_2019, hendrycks_2019}. It is well known that the latter scenarios impose a significant challenge for current practice, attributed to a common culprit referred to as overconfidence \citep{matan_90}. Specifically, popular discriminative approaches like SVMs~\citep{scheirer_2013,scheirer_2014} and neural networks~\citep{nguyen_2015, amodei_2016, guo_2017} chronically assign probabilities close to unity to their predictions, even when a category does not yet exist in the present model.

\begin{figure}
    \centering
    \input{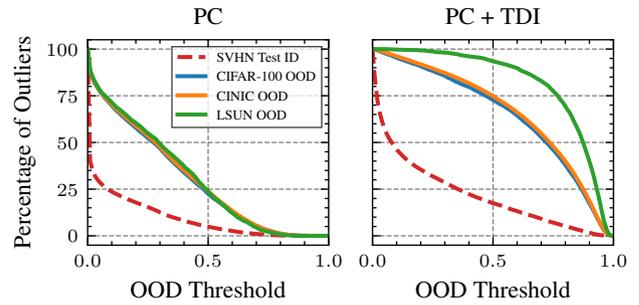}
    \caption{Conventional PCs are incapable of identifying OOD datasets (solid colors, outlier percentage should be large) while retaining correct predictions for ID data (dashed red, outlier percentage needs to be low to avoid rejection) independently of the decision threshold on the model's predictive entropy. Whereas their OOD detection precision drops rapidly with increasing threshold, TDI successfully distinguishes ID from OOD data across a wide range.}
    \label{fig:precision-ood}
\end{figure}

Unfortunately, the above challenge is not limited to discriminative models and
has recently resurfaced in the context of generative models. Various works
\citep{nalisnick_2019, ovadia_2019, mundt_2022} have empirically demonstrated
that different deep models, such as variational
auto-encoders~\citep{kingma2014auto} or normalizing flows~\citep{kobyzev2020normalizing, papamakarios2021normalizing}, have
analogous difficulties in separating data from arbitrary distributions from
those observed during training. Intuitively speaking, these models ``\emph{don't know what they don't know}''. In this paper, we show that a fairly new
family of generative models, probabilistic circuits (PCs)~\citep{choi_2020},
suffer from the same fate. Until now, these models have been generally assumed
to overcome the overconfidence problem faced by their deep neural network
counterparts, see e.g.~\citep{peharz_2020_icml}, ascribed primarily to PCs' ability for
\emph{tractable} and \emph{exact} inference. This assumption should
clearly be challenged, as highlighted by our empirical evidence in the left
panel of \cref{fig:precision-ood}. Here, we show the percentage of samples recognized as
outliers based on the predictive entropy $\sum_c p\left(y_c|x\right)\log p\left(y_c|x\right)$, with labels $y$, classes $c$, and data $x$, 
obtained by a PC in a classification scenario trained on an in-distribution (ID) dataset and tested on several out-of-distribution (OOD) datasets. When correctly identifying e.g. 95\% of SVHN as ID, the PC is only able to detect 24\% of the LSUN dataset successfully as OOD.

Inspired by the neural network literature, we consequently posit that PCs' inability to successfully recognize OOD instances is due to a lack of uncertainty quantification in their original formulation. More specifically, the ability to gauge the uncertainty in the model's parameters (also called \emph{epistemic} uncertainty) \citep{kendall_2017, hullermeier_2021} is required to indicate when the output is expected to be volatile and should thus not be trusted. In Bayesian neural networks~\citep{mackay_1992}, such uncertainty is achieved by placing a distribution on the weights and capturing the observed variation given some data. In arbitrary deep networks, the popular Monte Carlo dropout (MCD) \citep{gal_2016} measures uncertainty by conducting stochastic forward passes through a model with dropout \citep{srivastava_2014}, as a practical approximation based on Bernoulli distributions on the parameters.

In our work, we draw inspiration from the MCD approach and build upon its
success to quantify the uncertainty in PCs. However, in a crucial difference to neural networks and as the key
contribution of our paper, we
derive a closed-form solution to MCD leveraging the PC structure's clear probabilistic semantics. We refer to the derived procedure as \emph{tractable dropout inference} (TDI), which provides tractable uncertainty quantification in an efficient single forward pass by means of variance propagation. The right panel of \cref{fig:precision-ood} highlights how TDI successfully alleviates the overconfidence challenge of PCs and remarkably improves the OOD detection precision over the entire range of threshold values.
Measured over all OOD detection thresholds, TDI improves OOD precision over PCs without TDI by 2.2$\times$ on CIFAR-100 and CINIC, and 2.7$\times$ on LSUN (see \cref{sec:experimental_evaluation} for details). In summary, our key contributions are:
\begin{enumerate}
    \item We show that PCs suffer from overconfidence and fall short in distinguishing ID from OOD data.
    \item We introduce TDI, a novel inference routine that provides tractable uncertainty estimation. For this purpose, we derive a sampling-free analytical solution to MCD in PCs via variance propagation.
    \item We demonstrate TDI's robustness to distribution shifts and OOD data in three key scenarios with OOD data of different datasets, perturbed, and corrupted instances.
\end{enumerate}

\section{Related Work}
\label{sec:related_work}
The primary purpose and goal of our work is to introduce tractable uncertainty quantification to PCs to effectively perform OOD detection, an important aspect that has previously received little to no attention.
{On a broader scale, the current literature reflects minimal endeavors targeted at enhancing the robustness of PCs. These works
undertake the challenge from different perspectives.
For instance, to tackle overconfidence on in-domain misclassified samples when data is scarce,
\cite{maua_2017} introduced costly inference routines based
on credal sets that forgo tractable and general inference.
Recently, \cite{peddi_2022} presented a likelihood loss to involve artificially
perturbed samples during optimization. Although sharing the common aspect of challenging the robustness of PCs,
none of these works has provided an efficient tractable uncertainty quantification method to make arbitrary PCs more robust to different OOD
data without altering the model's parameters or compromising inference tractability. Thus, our most closely related works reside} in the respective area of uncertainty quantification, in particular, the imminently related approximations made in neural network counterparts \citep{blundell_2015, gal_2016, kendall_2017}. Gauging such model uncertainty in turn provides substantial value in various tasks, including OOD detection, robustness to corrupt and perturbed data, and several downstream applications. We provide a brief overview of the latter for the purpose of completeness, before concentrating on the essence of our paper in terms of immediately related methodology to estimate model uncertainty.

\textbf{OOD Detection and Use of Uncertainty:}
Uncertainty quantification based on Bayesian methods provides a theoretical foundation to assess when a model lacks
confidence in its parameters and predictions. Alternatively, several other directions have been proposed to deal 
with OOD inputs in the inference phase. Notably, several works across the decades have proposed to include 
various forms of reject options in classification.
These methods are often criticized for their lack of theoretical grounding, leading to a separate thread advocating 
for the challenge to be addressed through open set recognition.
We point to the recent review of \cite{boult_2019} for an overview of techniques. In a similar spirit,
assessment of uncertainty has been shown to be foundational in the application to
e.g. active learning \citep{gal_2017} or continual learning \citep{ebrahimi_2020}.
We emphasize that these techniques and applications are complementary to our work and are yet to be explored in PCs.
Similarly to prior neural network-based efforts of \cite{ovadia_2019} and \cite{nalisnick_2019}, we first show
that PCs are incapable of inherent OOD detection, before leaning on uncertainty to overcome the challenge with our TDI.

\textbf{Uncertainty Quantification:}
In a simplified picture, methods to estimate uncertainty could be attributed to two main categories: methods falling into a Bayesian framework and alternative non-Bayesian ones. As the categorization suggests, the latter do not ground their principle in Bayesian statistics and provide quantification in different forms, such as the size of a prediction interval or a score ~\citep{osband_2021, yu_2022_uai}. In contrast, Bayesian methods rely on a solid theoretical ground that allows for a clear interpretation. When applied to computation graphs like neural networks and PCs, the key
concept is to have a probability distribution over the parameters, in this
context, the weights of the graph. More formally, the model parameterization is framed as picking the parameters $\bm{\theta}$ (subject to optimization) from a prior probability distribution $p(\bm{\theta})$. We are then interested in the parameter configuration that most likely represents the data $\data$, i.e.,
$\argmax_{\bm{\theta}} p(\bm{\theta}|\data)$. To account for model uncertainty, it would be necessary to integrate over the parameters, which is intractable for many models.

A considerable number of works have pursued this direction for
deep neural architectures. Bayesian neural networks~\citep{mackay_1992, neal_2012} have
initially paved the way to model uncertainty, but given the immense computational cost, alternatives focus on several
cheap approximations. Popular ways are to back-propagate uncertainty through
the gradients~\citep{blundell_2015, hernandez_lobato_2015, mishkin_2018, maddox_2019}, make use of variational inference~\citep{graves_2011, louizos_2016} or draw connections to Gaussian Processes ~\citep{gal_2016, khan_2019}.
A related approximate approach is based on ensembles, where the underlying idea is to relate the
model uncertainty with the statistics computed over the various ensemble components. To obtain uncertainty estimates, most of these
approaches need to train multiple ensemble components~\citep{hansen_1990, lakshminarayanan_2017} or larger
overparameterized singletons and treat them as an ensemble of
subnetworks~\citep{antoran_2020, daxberger_2021}.
Among the Bayesian methods for learning a PC on propositional knowledge bases, \cite{cerutti_2022} deal with uncertainty estimation for conditional boolean queries by attaching a second circuit to the PC. 

\textbf{Monte Carlo Dropout:}
The natural question for Bayesian methods is how to sample
the parameters $\bm{\theta}$ from the posterior $p\left(\bm{\theta}\cbar\data\right)$, taking the
high-dimensional and highly non-convex nature of the probability distribution for complex networks into account, which leads to intractable standard
sampling methods~\citep{izmailov_2021}. \cite{gal_2016}
have reframed dropout \citep{srivastava_2014} as a Bayesian approximation to assess model uncertainty. Originally, dropout is a method proposed to avoid overfitting and
improve generalization by including a stochastic chance $p$ of removing a connection between units of an adjacent layer. \cite{gal_2016}'s key realization is that dropout allows to cheaply sample from the posterior under the assumption of a Bernoulli distribution on the weights.
In essence, MCD approximates the integration over the parameters with a summation over a finite set of $n$ drawn sets of parameters
$\bm{\theta}_i \sim p\left(\bm{\theta}\cbar\data\right)$. By using the set of $n$ predicted
values, the first and the second raw moments can be computed. The former is then used as the prediction and the latter as an estimate of model uncertainty.

The essential advantage of MCD is its simple applicability, which has led to a wide range of immediate applications \citep{kendall_2017, gal_2017, kendall2017a, miller_2018}. In our work, we draw inspiration from MCD and its vast impact. However, instead of approximating the uncertainty with a Monte Carlo simulation in PCs, we perform variance propagation from the leaf to the root nodes with a single pass, with which we derive a sampling-free, closed-form solution to model uncertainty.

\section{Tractable Dropout Inference}
\label{sec:tdi}
In this section, we first introduce preliminaries with respect to PCs, before continuing to delve into a step-by-step derivation of how to obtain sampling-free uncertainties with TDI.

\subsection{Preface: Probabilistic Circuits}
\label{sec:tdi:pcs}

In this work, we refer to a relevant class of PCs, i.e.,
sum-product networks (SPNs)~\citep{poon_2011}. In the family of
tractable probabilistic models, SPNs stand out for their inference capabilities
and great representational power~\citep{delalleau_2011}.
They hold important structural properties
such as \emph{smoothness} and \emph{decomposability}
that enable the efficient encoding of a valid
probability distribution. In the following, we first formally introduce SPNs and their important properties.

\begin{figure*}
  \centering
  \begin{subfigure}[t]{0.333\linewidth}
\centering \scalebox{0.3333}{
    \begin{tikzpicture}[minimum size=8mm, inner sep=0pt,
      align=center,
prod/.style={circle,tab10blue, draw, fill=tab10blue!10,line width=\nodethickness, minimum size=\minnodesize, path picture={
    \draw[tab10blue]
    (path picture bounding box.south east) -- (path picture bounding box.north west) (path picture bounding box.south west) -- (path picture bounding box.north east);
    }},
    sum/.style={circle, tab10green, draw, fill=tab10green!10, line width=\nodethickness, minimum size=\minnodesize, path picture={
    \draw[tab10green]
    (path picture bounding box.south) -- (path picture bounding box.north) (path picture bounding box.west) -- (path picture bounding box.east);
    }},
    leaf/.style={circle, draw, line width=\nodethickness, minimum
      size=\minnodesize},
    pcedge/.style={thick,{Stealth[scale=1.25]}-, font=\fontsize{16}{0}\selectfont},
    ]

    \node[sum, label={[font=\fontsize{16}{0}\selectfont]north:$\spn$}] (s1) {};

    \node[prod] (p11) [below left=\nodedist and 1.25\nodedist of s1] {};
    \node[prod] (p12) [below right=\nodedist and 1.25\nodedist of s1] {};

    \node[sum] (s21) [below left=\prodnodedist and 0.75\prodnodedist of p11] {};
    \node[leaf] (l22) [below right=\prodnodedist and 0.75\prodnodedist of p11] {\Large $\X_{3}$};
    \node[leaf] (l23) [below left=\prodnodedist and 0.75\prodnodedist of p12] {\Large $\X_{1}$};
    \node[sum] (s24) [below right=\prodnodedist and 0.75\prodnodedist of p12] {};

    \node[prod] (p21) [below left=\nodedist and 0.5\nodedist of s21] {};
    \node[prod] (p22) [below right=\nodedist and 1.25\nodedist of s21] {};
    \node[prod] (p23) [below left=\nodedist and 1.25\nodedist of s24] {};
    \node[prod] (p24) [below right=\nodedist and 0.5\nodedist of s24] {};

    \node[leaf] (l31) [below left=1.25\nodedist and 0.3\leafnodedist of p21] {\Large $\X_{1}$};
    \node[leaf] (l32) [below right=1.25\nodedist and 0.3\leafnodedist of p21] {\Large $\X_{2}$};
    \node[leaf] (l33) [below left=1.25\nodedist and 0.3\leafnodedist of p22] {\Large $\X_{1}$};
    \node[leaf] (l34) [below right=1.25\nodedist and 0.3\leafnodedist of p22] {\Large $\X_{2}$};
    \node[leaf] (l35) [below left=1.25\nodedist and 0.3\leafnodedist of p23] {\Large $\X_{2}$};
    \node[leaf] (l36) [below right=1.25\nodedist and 0.3\leafnodedist of p23] {\Large $\X_{3}$};
    \node[leaf] (l37) [below left=1.25\nodedist and 0.3\leafnodedist of p24] {\Large $\X_{2}$};
    \node[leaf] (l38) [below right=1.25\nodedist and 0.3\leafnodedist of p24] {\Large $\X_{3}$};

    \draw[pcedge] (s1) -- (p11) node[midway,fill=white] {$\prodnode_{1}$};
    \draw[pcedge] (s1) -- (p12) node[midway,fill=white] {$\prodnode_{2}$};

    \draw[pcedge] (p11) -- (s21) node[midway,fill=white] {$\sumnode_{2}$};
    \draw[pcedge] (p11) -- (l22) node[midway,fill=white] {$\leafnode_{1}^{3}$};
    \draw[pcedge] (p12) -- (l23) node[midway,fill=white] {$\leafnode_{2}^{1}$};
    \draw[pcedge] (p12) -- (s24) node[midway,fill=white] {$\sumnode_{3}$};

    \draw[pcedge] (s21) -- (p21) node[midway,fill=white] {$\prodnode_{3}$};
    \draw[pcedge] (s21) -- (p22) node[midway,fill=white] {$\prodnode_{4}$};
    \draw[pcedge] (s24) -- (p23) node[midway,fill=white] {$\prodnode_{5}$};
    \draw[pcedge] (s24) -- (p24) node[midway,fill=white] {$\prodnode_{6}$};

    \draw[pcedge] (p21) -- (l31) node[midway,fill=white] {$\leafnode_{3}^{1}$};
    \draw[pcedge] (p21) -- (l32) node[midway,fill=white] {$\leafnode_{4}^{2}$};
    \draw[pcedge] (p22) -- (l33) node[midway,fill=white] {$\leafnode_{5}^{1}$};
    \draw[pcedge] (p22) -- (l34) node[midway,fill=white] {$\leafnode_{6}^{2}$};
    \draw[pcedge] (p23) -- (l35) node[midway,fill=white] {$\leafnode_{7}^{2}$};
    \draw[pcedge] (p23) -- (l36) node[midway,fill=white] {$\leafnode_{8}^{3}$};
    \draw[pcedge] (p24) -- (l37) node[midway,fill=white] {$\leafnode_{9}^{2}$};
    \draw[pcedge] (p24) -- (l38) node[midway,fill=white] {$\leafnode_{10}^{3}$};
  \end{tikzpicture}
} 
    \caption{Probabilistic inference}
    \label{fig:graphs:pc}
  \end{subfigure}%
  \begin{subfigure}[t]{0.333\linewidth}
\centering \scalebox{0.3333}{
    \begin{tikzpicture}[minimum size=8mm, inner sep=0pt,
      align=center,
    prod/.style={circle,tab10blue, draw, fill=tab10blue!10,line width=\nodethickness, minimum size=\minnodesize, path picture={
    \draw[tab10blue]
    (path picture bounding box.south east) -- (path picture bounding box.north west) (path picture bounding box.south west) -- (path picture bounding box.north east);
    }},
    proddrop/.style={circle,tab10blue!30, draw, fill=tab10blue!05,line width=\nodethickness, minimum size=\minnodesize, path picture={
    \draw[tab10blue!30]
    (path picture bounding box.south east) -- (path picture bounding box.north west) (path picture bounding box.south west) -- (path picture bounding box.north east);
    }},
    sum/.style={circle, tab10green, draw, fill=tab10green!10, line width=\nodethickness, minimum size=\minnodesize, path picture={
    \draw[tab10green]
    (path picture bounding box.south) -- (path picture bounding box.north) (path picture bounding box.west) -- (path picture bounding box.east);
    }},
    leafdrop/.style={circle, gray!30, draw, line width=\nodethickness, minimum size=\minnodesize},
    leaf/.style={circle, draw, line width=\nodethickness, minimum
      size=\minnodesize},
    pcedge/.style={thick,{Stealth[scale=1.25]}-, font=\fontsize{16}{0}\selectfont},
    pcedgedrop/.style={thick,{Stealth[scale=1.25]}-, gray!50, font=\fontsize{16}{0}\selectfont},
    ]

    \node[sum, label={[font=\fontsize{16}{0}\selectfont]north:$\spn$}] (s1) {};

    \node[prod] (p11) [below left=\nodedist and 1.25\nodedist of s1] {};
    \node[prod] (p12) [below right=\nodedist and 1.25\nodedist of s1] {};

    \node[sum] (s21) [below left=\prodnodedist and 0.75\prodnodedist of p11] {};
    \node[leaf] (l22) [below right=\prodnodedist and 0.75\prodnodedist of p11] {\Large $\X_{3}$};
    \node[leaf] (l23) [below left=\prodnodedist and 0.75\prodnodedist of p12] {\Large $\X_{1}$};
    \node[sum] (s24) [below right=\prodnodedist and 0.75\prodnodedist of p12] {};

    \node[proddrop] (p21) [below left=\nodedist and 0.5\nodedist of s21] {};
    \node[prod] (p22) [below right=\nodedist and 1.25\nodedist of s21] {};
    \node[prod] (p23) [below left=\nodedist and 1.25\nodedist of s24] {};
    \node[proddrop] (p24) [below right=\nodedist and 0.5\nodedist of s24] {};

    \node[leafdrop] (l31) [below left=1.25\nodedist and 0.3\leafnodedist of p21] {\Large $\X_{1}$};
    \node[leafdrop] (l32) [below right=1.25\nodedist and 0.3\leafnodedist of p21] {\Large $\X_{2}$};
    \node[leaf] (l33) [below left=1.25\nodedist and 0.3\leafnodedist of p22] {\Large $\X_{1}$};
    \node[leaf] (l34) [below right=1.25\nodedist and 0.3\leafnodedist of p22] {\Large $\X_{2}$};
    \node[leaf] (l35) [below left=1.25\nodedist and 0.3\leafnodedist of p23] {\Large $\X_{2}$};
    \node[leaf] (l36) [below right=1.25\nodedist and 0.3\leafnodedist of p23] {\Large $\X_{3}$};
    \node[leafdrop] (l37) [below left=1.25\nodedist and 0.3\leafnodedist of p24] {\Large $\X_{2}$};
    \node[leafdrop] (l38) [below right=1.25\nodedist and 0.3\leafnodedist of p24] {\Large $\X_{3}$};

    \draw[pcedge] (s1) -- (p11) node[midway,fill=white] {$\prodnodeD_{1}$};
    \draw[pcedge] (s1) -- (p12) node[midway,fill=white] {$\prodnodeD_{2}$};

    \draw[pcedge] (p11) -- (s21) node[midway,fill=white] {$\sumnodeD_{2}$};
    \draw[pcedge] (p11) -- (l22) node[midway,fill=white] {$\leafnodeD_{1}^{3}$};
    \draw[pcedge] (p12) -- (l23) node[midway,fill=white] {$\leafnodeD_{2}^{1}$};
    \draw[pcedge] (p12) -- (s24) node[midway,fill=white] {$\sumnodeD_{3}$};

    \draw[pcedgedrop] (s21) -- (p21) node[midway,fill=white] {$\prodnodeD_{3}$};
    \draw[pcedge] (s21) -- (p22) node[midway,fill=white] {$\prodnodeD_{4}$};
    \draw[pcedge] (s24) -- (p23) node[midway,fill=white] {$\prodnodeD_{5}$};
    \draw[pcedgedrop] (s24) -- (p24) node[midway,fill=white] {$\prodnodeD_{6}$};

    \draw[pcedgedrop] (p21) -- (l31) node[midway,fill=white] {$\leafnodeD_{3}^{1}$};
    \draw[pcedgedrop] (p21) -- (l32) node[midway,fill=white] {$\leafnodeD_{4}^{2}$};
    \draw[pcedge] (p22) -- (l33) node[midway,fill=white] {$\leafnodeD_{5}^{1}$};
    \draw[pcedge] (p22) -- (l34) node[midway,fill=white] {$\leafnodeD_{6}^{2}$};
    \draw[pcedge] (p23) -- (l35) node[midway,fill=white] {$\leafnodeD_{7}^{2}$};
    \draw[pcedge] (p23) -- (l36) node[midway,fill=white] {$\leafnodeD_{8}^{3}$};
    \draw[pcedgedrop] (p24) -- (l37) node[midway,fill=white] {$\leafnodeD_{9}^{2}$};
    \draw[pcedgedrop] (p24) -- (l38) node[midway,fill=white] {$\leafnodeD_{10}^{3}$};
  \end{tikzpicture}
}
    \caption{Monte Carlo dropout inference}
    \label{fig:graphs:mcd}
  \end{subfigure}%
  \begin{subfigure}[t]{0.333\linewidth}
\newcommand{\prodlabel}[1]{$\Varinline{\prodnodeD_{#1}}$}
\newcommand{\sumlabel}[1]{$\Varinline{\sumnodeD_{#1}}$}
\newcommand{\leaflabel}[2]{$\Varinline{\leafnodeD_{#1}^{#2}}$}
\centering \scalebox{0.3333}{
    \begin{tikzpicture}[minimum size=8mm, inner sep=0pt,
      align=center,
prod/.style={circle,tab10blue, draw, fill=tab10blue!10,line width=\nodethickness, minimum size=\minnodesize, path picture={
    \draw[tab10blue]
    (path picture bounding box.south east) -- (path picture bounding box.north west) (path picture bounding box.south west) -- (path picture bounding box.north east);
    }},
    sum/.style={circle, tab10green, draw, fill=tab10green!10, line width=\nodethickness, minimum size=\minnodesize, path picture={
    \draw[tab10green]
    (path picture bounding box.south) -- (path picture bounding box.north) (path picture bounding box.west) -- (path picture bounding box.east);
    }},
    leaf/.style={circle, draw, line width=\nodethickness, minimum
      size=\minnodesize},
    pcedge/.style={thick,{Stealth[scale=1.25]}-, font=\fontsize{16}{0}\selectfont},
    ]

    \node[sum, label={[font=\fontsize{16}{0}\selectfont]north:$\Varinline{\spn}$}] (s1) {};

    \node[prod] (p11) [below left=\nodedist and 1.25\nodedist of s1] {};
    \node[prod] (p12) [below right=\nodedist and 1.25\nodedist of s1] {};

    \node[sum] (s21) [below left=\prodnodedist and 0.75\prodnodedist of p11] {};
    \node[leaf] (l22) [below right=\prodnodedist and 0.75\prodnodedist of p11] {\Large $\X_{3}$};
    \node[leaf] (l23) [below left=\prodnodedist and 0.75\prodnodedist of p12] {\Large $\X_{1}$};
    \node[sum] (s24) [below right=\prodnodedist and 0.75\prodnodedist of p12] {};

    \node[prod] (p21) [below left=\nodedist and 0.5\nodedist of s21] {};
    \node[prod] (p22) [below right=\nodedist and 1.25\nodedist of s21] {};
    \node[prod] (p23) [below left=\nodedist and 1.25\nodedist of s24] {};
    \node[prod] (p24) [below right=\nodedist and 0.5\nodedist of s24] {};

    \node[leaf] (l31) [below left=1.25\nodedist and 0.3\leafnodedist of p21] {\Large $\X_{1}$};
    \node[leaf] (l32) [below right=1.25\nodedist and 0.3\leafnodedist of p21] {\Large $\X_{2}$};
    \node[leaf] (l33) [below left=1.25\nodedist and 0.3\leafnodedist of p22] {\Large $\X_{1}$};
    \node[leaf] (l34) [below right=1.25\nodedist and 0.3\leafnodedist of p22] {\Large $\X_{2}$};
    \node[leaf] (l35) [below left=1.25\nodedist and 0.3\leafnodedist of p23] {\Large $\X_{2}$};
    \node[leaf] (l36) [below right=1.25\nodedist and 0.3\leafnodedist of p23] {\Large $\X_{3}$};
    \node[leaf] (l37) [below left=1.25\nodedist and 0.3\leafnodedist of p24] {\Large $\X_{2}$};
    \node[leaf] (l38) [below right=1.25\nodedist and 0.3\leafnodedist of p24] {\Large $\X_{3}$};

    \draw[pcedge] (s1) -- (p11) node[midway,fill=white] {\prodlabel{1}};
    \draw[pcedge] (s1) -- (p12) node[midway,fill=white] {\prodlabel{2}};

    \draw[pcedge] (p11) -- (s21) node[midway,fill=white] {\sumlabel{2}};
    \draw[pcedge] (p11) -- (l22) node[midway,fill=white] {\leaflabel{1}{3}};
    \draw[pcedge] (p12) -- (l23) node[midway,fill=white] {\leaflabel{2}{1}};
    \draw[pcedge] (p12) -- (s24) node[midway,fill=white] {\sumlabel{3}};

    \draw[pcedge] (s21) -- (p21) node[midway,fill=white] {\prodlabel{3}};
    \draw[pcedge] (s21) -- (p22) node[midway,fill=white] {\prodlabel{4}};
    \draw[pcedge] (s24) -- (p23) node[midway,fill=white] {\prodlabel{5}};
    \draw[pcedge] (s24) -- (p24) node[midway,fill=white] {\prodlabel{6}};

    \draw[pcedge] (p21) -- (l31) node[midway,fill=white, xshift=-0.3cm] {\leaflabel{3}{1}};
    \draw[pcedge] (p21) -- (l32) node[midway,fill=white, xshift=0.3cm] {\leaflabel{4}{2}};
    \draw[pcedge] (p22) -- (l33) node[midway,fill=white, xshift=-0.3cm] {\leaflabel{5}{1}};
    \draw[pcedge] (p22) -- (l34) node[midway,fill=white, xshift=0.3cm] {\leaflabel{6}{2}};
    \draw[pcedge] (p23) -- (l35) node[midway,fill=white, xshift=-0.3cm] {\leaflabel{7}{2}};
    \draw[pcedge] (p23) -- (l36) node[midway,fill=white, xshift=0.3cm] {\leaflabel{8}{3}};
    \draw[pcedge] (p24) -- (l37) node[midway,fill=white, xshift=-0.3cm] {\leaflabel{9}{2}};
    \draw[pcedge] (p24) -- (l38) node[midway,fill=white, xshift=0.3cm] {\leaflabel{10}{3}};
  \end{tikzpicture}
}
    \caption{Tractable dropout inference}
    \label{fig:graphs:cf}
  \end{subfigure}%
  \caption{Forward pass illustration of a PC (\subref{fig:graphs:pc}), a PC with
    MCD sampling (\subref{fig:graphs:mcd}), and a PC with TDI based on variance propagation through the graph (\subref{fig:graphs:cf}). Whereas a single MCD forward pass only simulates one instantiation of a possible
    subgraph, the PC with TDI directly propagates the variances through the graph in a
    single pass (expectations and covariances are left out for visual clarity in the illustration, see \cref{sec:tdi:derivation} for full equations).}
  \label{fig:graphs}
\end{figure*}
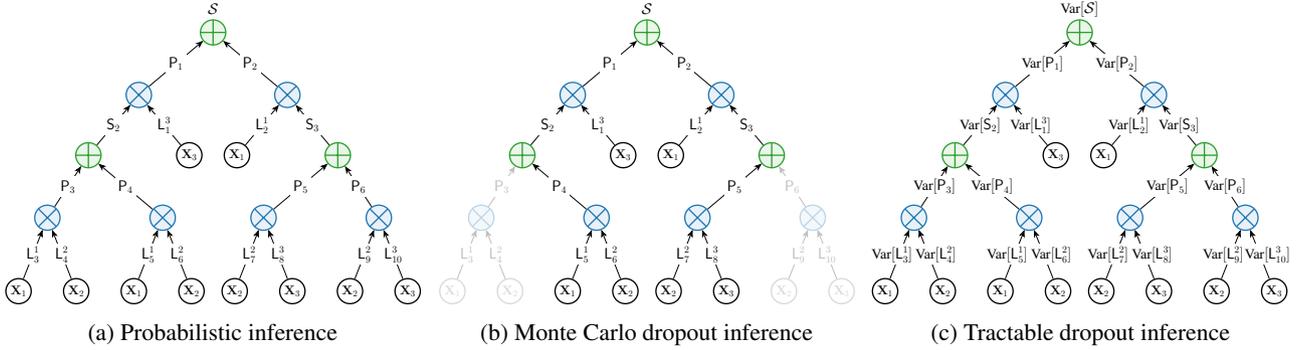

\textbf{Definition:}
An SPN $\spn$ is a computational graph defined by a rooted directed acyclic graph (DAG), encoding
a probability distribution $p_{\X}$ over a set of random variables (RVs)
$\X = \{X_1, \ldots, X_n\}$, where inner nodes are either sum nodes $\sumnode$ or product
nodes $\prodnode$ over their children,
 and leaves $\leafnode$ are valid probability distributions defined on a
subset of the RVs $\Z \subseteq \X$. Each node $\node \in \spn$ has a \emph{scope},
$\scope{\node} \subseteq \X$, defined as the set of RVs appearing in
its descendant leaves. Each edge $(\sumnode_{i} \rightarrow \node_{j})$ connecting a sum node
$\sumnode_i$ to one of its children $\node_j \in \ch\left(\sumnode_i\right)$ has a non-negative weight $w_{ij}$, with
$\sum_j w_{ij} = 1$. Sum nodes represent mixtures over the probability
distributions encoded by their children, while product nodes represent
factorizations over contextually independent distributions. In summary, an SPN
can be viewed as a deep hierarchical mixture model of different factorizations.
An illustration of this kind of PC is shown in \cref{fig:graphs:pc}.

To encode a valid probability distribution, an SPN has to fulfill
two structural requirements~\citep{poon_2011}. One is
\textit{decomposability}, i.e., the scopes of the children of each product node
need to be disjoint, which allows distributing the NP-hard computation of
integrals (e.g. the partition function), to leaves where we require
this computation to be tractable. This condition can be fulfilled, either by an explicit form, e.g., from an exponential family distribution, or by the architectural design of the leaf
density estimators. The second requirement is \textit{smoothness}, constraining the scopes of
the children of each sum node to be identical (this is also referred
to as \emph{completeness}). This constraint is important to encode a
valid distribution that does not over- or underestimate some RVs states. In a
valid SPN, the probability assigned to a given state $\x$ of the RVs
$\X$ is given by the root node and will be denoted as $\spn(\x) = p_{\X}(\X = \x)$.

\textbf{Tractable Inference:}
Given an SPN $\spn$, $\spn\!\left(\x\right)$ is computed
by evaluating the network bottom-up. When evaluating a leaf node $\leafnode_{i}$
with scope $X_j$, $\leafnode_{i}\!\left(x_j\right)$ corresponds to the probability of the
state $x_j$. The value of a product node $\prodnode_{i}$
corresponds to the product of its children's values:
$\prodnode_{i}\!\left(\x_{|\scope{\prodnode_{i}}}\right)=\prod_{\node_{j} \in \ch\left(\prodnode_i\right)} \node_{j}(\x_{|\scope{\node_{j}}})$.
The value of a sum node $\sumnode_{i}$ is computed as the weighted sum
of its children's values:
$\sumnode_{i}\left(\x_{|\scope{\sumnode_{i}}}\right) = \sum_{\node_{j} \in \ch\left(\sumnode_i\right)}w_{ij} \node_{j}(\x_{|\scope{\node_{j}}})$.
All the exact marginal and conditional probabilities, also with different amount
of evidence, the exact partition function, and even approximate most probable explanation and maximum a posteriori
states can be computed in time and space linear in the network's \emph{size}, i.e. number of edges~\citep{poon_2011, peharz_2015}.

Being part of the PC family, SPNs share the basic building
blocks and procedures when performing inference with other classes of PCs. Thus, the contributions
that we introduce in this paper can also be easily applied to other
models of the PC family~\citep{choi_2020}. We further note that because TDI is presently framed as an inference routine, training follows conventional algorithms. Whereas the inclusion of uncertainties to guide training itself is certainly intriguing, we defer this prospect to future work and concentrate on the means to quantify uncertainty to overcome overconfidence and detect OOD data.  

\subsection{Deriving TDI}
\label{sec:tdi:derivation}
One of the crucial aspects in which PCs stand out compared to neural networks is
that they have clear probabilistic semantics. Leaf nodes are normalized
tractable distributions and sum nodes represent weighted
mixtures over different factorizations, encoded by the product nodes, on the
same scope. When dropout is applied to PCs, it thus entails an easier
interpretation. At first look, similar to the case for neural networks, MCD
at sum nodes would perform a sort of model averaging between a randomly selected set
of models (sub-graphs). Naively, we could follow the same procedure as proposed
in \cite{gal_2016} and conduct stochastic forward passes with a more intuitive
interpretation on mixtures over different factorizations (compared to a
black-box neural network); see \cref{fig:graphs:mcd} for an illustration. The
variance of the different models is then interpreted as the uncertainty w.r.t. a
specific input. However, we will now show that it is also possible to derive TDI as a
tractable uncertainty estimate in a single forward pass in PCs. With our closed-form derivations, we thus provide an
analytical solution to this measure of uncertainty.

\textbf{Core Idea -- Sampling-free Uncertainty:}
The general idea is to derive closed-form expressions of the expectation,
variance, and covariance for sum and product nodes as a function of their children.
The model uncertainty at the root nodes is then recursively computed by
performing variance propagation from the leaf nodes to the root nodes
in a single bottom-up pass through the graph structure (see \cref{fig:graphs:cf}). This procedure results in
uncertainty estimates without sampling from the graph multiple times, as is required by MCD.

In TDI we start by viewing sum nodes as linear combinations of RVs over their Bernoulli
dropout RVs and their children:
\begin{align}
  \label{eq:sum-prod-definitions}
  \sumnodeD = \sum_{i} \delta_{i} w_{i} \prodnode_{i} \, ,
\end{align}
where $\delta_{i} \sim \text{Bern}\!\left(q\right)$ and $p = 1-q$ corresponds to
the dropout probability.

We will now provide the expectation, variance, and covariance closed-form
solutions for sum, product, and leaf nodes.
Here, we present the most relevant derivations
for these quantities. For full derivations and theoretical details, we refer to Appendix A.

\subsubsection{Expectation: The Point Estimate}
\label{sec:tdi:derivation:expectation}
Using the linearity of the expectation, we push the expectation into the sum
and make use of the independence between the child nodes $\nodeD_{i}$ and the
Bernoulli RVs $\delta_{i}$ to extract $\Einline{\delta_i} = q$:
\begin{align}
  \label{eq:sum-expectation}
  \E{\sumnodeD} = q \sum_{i} w_{i}\E{\nodeD_{i}} \, .
\end{align}

The decomposability of product nodes ensures the independence of their children
w.r.t. each other, which leads to the product node expectation simply becoming the
product over the expectations of its children, i.e.,
\begin{align}
  \label{eq:prod-expectation}
  \E{\prodnodeD} =  \prod_{i} \E{\nodeD_{i}} \, .
\end{align}

\subsubsection{Variance: The Uncertainty Proxy}
\label{sec:tdi:derivation:variance}
Similar to the original work in neural networks of~\cite{gal_2016},
we will use the variance as
the proxy for the uncertainty. The sum node variance decomposes
into a sum of two terms. The first term is based on the variances and
expectations of its children and the second term accounts for the covariance
between the combinations of all children, i.e.,
\begin{alignat}{2}
  \label{eq:sum-variance}
  \Var{\sumnodeD} &= && q\sum_{i} w_{i}^{2} \left( \Var{\nodeD_{i}} + p\E{\nodeD_{i}}^{2}  \right) \notag\\
                  &  &&+ q^{2}\sum_{i\neq j}w_{i}w_{j}\Cov{\nodeD_{i}, \nodeD_{j}} \, .
\end{alignat}
Analogously, the product node variance decomposes into two product terms. By
applying the product of independent variables rule, we obtain:
\begin{align}
  \label{eq:prod-variance}
  \Var{\prodnodeD} &= \prod_{i} \left( \Var{\nodeD_{i}} + \E{\nodeD_{i}}^{2} \right) - \prod_{i} \E{\nodeD_{i}}^{2} \, .
\end{align}
\subsubsection{Covariance: The Evil}
\label{sec:tdi:derivation:covariance}
The covariance between two sum nodes, $\sumnodeD^{A}$ and $\sumnodeD^{B}$,
neatly decomposes into a weighted sum of all covariance combinations of the sum
node children:
\begin{align}
  \label{eq:sum-covariance}
  \text{Cov}\!\left[ \sumnodeD^{A}, \sumnodeD^{B}\right] = q^2\sum_{i}w_{i}^{A}\sum_{j}w_{j}^{B}\Cov{\nodeD_{i}^{A},\node_{j}^{B}} \, .
\end{align}
For an arbitrary graph, we are unable to provide a closed-form solution of the covariance between two product
nodes due to the first expectation in the product node
covariance:
\begin{alignat}{2}
  \Cov{\prodnodeD^{A},\prodnodeD^{B}} &= &&\E{\prod_{i}\nodeD_{i}^{A} \prod_j \nodeD_{j}^{B}} \notag\\
                                      & &&- \prod_{i}\E{\nodeD_{i}^{A}} \prod_j \E{\nodeD_{j}^{B}} \, , \label{eq:prod-covariance}
\end{alignat}
which cannot be simplified without any structural knowledge about independencies
between the children of $\nodeD^{A}$ and $\nodeD^{B}$, as they may share
a common subset of nodes, deeper down in the PC structure. Fortunately, we explore
three possible solutions to solve \cref{eq:prod-covariance} in the following, of which variants b) and c) are always applicable.

\textbf{a) Structural Knowledge:}
To simplify \cref{eq:prod-covariance} we can easily exploit structural knowledge
of the DAG. The simplest solution is a structure in which we know that two
product nodes $\prodnodeD^{A}$ and $\prodnodeD^{B}$ are not common ancestors of
any node, resulting in the independence $\prodnodeD^{A} \ind \prodnodeD^{B}$ and thus $\Covinline{\prodnodeD^{A}, \prodnodeD^{B}} = 0$. This
constraint is always given in tree-structured PCs. Because simple in practice,
tree structures are generated by the most common structure learner for SPNs such
as LearnSPN~\citep{gens_2013}, ID-SPN~\citep{rooshenas_2014}, and
SVD-SPN~\citep{adel_2015}.

For binary tree random and tensorized (RAT) structures \citep{peharz_2020}, \cref{eq:prod-covariance} can be simplified to
\begin{alignat}{2}
  \Cov{\prodnodeD_{l,r}, \prodnodeD_{l',r'}} =&\, &&\Cov{\sumnodeD^L_{l}, \sumnodeD^L_{l'}} \E{\sumnodeD^R_{r}} \E{\sumnodeD^R_{r'}} \notag \\
  &+\, &&\Cov{\sumnodeD^R_{r}, \sumnodeD^R_{r'}} \E{\sumnodeD^L_{l}} \E{\sumnodeD^L_{l'}} \notag \\
  &+\, &&\Cov{\sumnodeD^L_{l}, \sumnodeD^L_{l'}} \Cov{\sumnodeD^R_{r}, \sumnodeD^R_{r'}} \, . \label{eq:cov-rat}
\end{alignat}
The covariance of two product nodes now only depends on the covariance of the
input sum nodes of the same graph partition ($L$ or $R$) for which we can
plug in \cref{eq:sum-covariance}.

\textbf{b) It's Somewhere in Here -- Covariance Bounds:}
Whereas knowledge about the specific PC structure can facilitate the covariance
computation, when not available, we can
alternatively obtain a lower and upper bound of the covariance, making use of
the Cauchy-Schwarz inequality:
\begin{alignat}{2}
  \label{eq:cauchy-schwarz}
  &\Cov{\ndi, \ndj}^{2} &&\leq \Var{\ndi}\Var{\ndj} \\
  \Leftrightarrow \quad &\Cov{\ndi, \ndj} &&\in \left[ - \sqrt{\Var{\ndi}\Var{\ndj}}, \right. \notag\\
  & && \left.+ \sqrt{\Var{\ndi}\Var{\ndj}} \right] \quad . \label{eq:cauchy-schwarz:last}
\end{alignat}

\textbf{c) The Copy-paste Solution:}
A third alternative to using structural knowledge or giving covariance
bounds is via a ``copy-paste'' augmentation of the DAG, that enforces the
covariance between two nodes, $\nodeD_{A}$ and $\nodeD_{B}$, to be zero
by treating their common child as two separate nodes. That
is, for each node $\nodeD_{C}$ where a
$\text{Path}_{A} := \nodeD_{A} \rightarrow \nodeD_{C}$ and a second
$\text{Path}_{B} := \nodeD_{B} \rightarrow \nodeD_{C}$ exists, we can ``copy''
$\nodeD_{C}$ to obtain an equivalent node $\nodeD_{C'}$ and replace the original
$\nodeD_{C}$ in $\text{Path}_{B}$ with the copy $\nodeD_{C'}$. With this simple
procedure, we can enforce a tree structure on the PC, resulting in the
covariance between two children of a node $\nodeD$ to be zero.
In practice, we do not need to modify the DAG. Instead, we can simply ignore the covariance
terms and thereby obtain this DAG transformation during the TDI procedure implicitly.

\subsubsection{Leaf Nodes}
\label{sec:tdi:derivation:leaf-nodes}

As leaf nodes are free of any dropout Bernoulli variables, their expectation,
variance, and covariance degrade to the leaf node value and zero respectively,
i.e.,
\begin{align}
  \label{eq:leaf-exp-var-cov}
  \E{\leafnodeD} = \leafnodeD, \quad \Var{\leafnodeD} = 0, \quad \Cov{\leafnodeD_{i},\leafnodeD_{j}} = 0 \, .
\end{align}
While the above is a valid choice, this framework
further allows including prior knowledge about aleatoric and epistemic
uncertainty, by setting $\Varinline{\leafnodeD} >0$ and
$\Covinline{\leafnodeD_{i}, \leafnodeD_{j}} \neq 0$. This additionally
highlights the advantage over the MCD procedure, where the inclusion of prior knowledge
is not possible.

\subsubsection{Classification Uncertainty}
\label{sec:tdi:derivation:classification}
For classification in PCs, we can express the class conditionals $p\left(\x \cbar y_i\right) = \sumnodeD_i$ as root nodes with class priors $p\left(y_i\right)=c_i$ and obtain the posterior  via Bayes' rule, i.e.,
\begin{align}
  \label{eq:1}
  p\left(y_{i} \cbar \x\right) = \frac{p\left(\x \cbar y_{i}\right) p\left(y_{i}\right)}{\sum_{j} p\left(\x \cbar y_{j}\right) p\left(y_{j}\right)} = \frac{\sumnodeD_{i}\ci}{\sum_{j}\sumnodeD_{j}\cj} \, .
\end{align}
In our case, the expectation and variance of the posterior are that of a random variable
ratio, $\E{\frac{A}{B}}$ and $\Var{\frac{A}{B}}$, with $A = \sumnodeD_{i}\ci$ and
$B = \sum_{j}\sumnodeD_{j}\cj$. This ratio is generally not well-defined, but can
be approximated with a second-order Taylor
approximation~\citep{ratio_seltman}: 
\begin{align}
  \E{\frac{A}{B}} &\approx \frac{\E{A}}{\E{B}} - \frac{\Cov{A, B}}{\left( \E{B} \right)^{2}} + \frac{\Var{B}\E{A}}{\left( \E{B} \right)^{3}} \label{eq:exp-rv-ratio}\\
  \Var{\frac{A}{B}} &\approx \frac{\E{A}^{2}}{\E{B}^{2}} \left[ \frac{\Var{A}}{\E{A}^{2}} - 2 \frac{\Cov{A, B}}{\E{A}\E{B}} + \frac{\Var{B}}{\E{B}^{2}} \right] . \label{eq:var-rv-ratio}
\end{align}
We will now resolve every component of \cref{eq:exp-rv-ratio,eq:var-rv-ratio}.
The expectations are straightforward:
\begin{align}
  \E{A} &= \E{\sumnodeD_{i}\ci} = \E{\sumnodeD_{i}}\ci \label{eq:exp-x}\\
  \E{B} &= \E{\sum_{j}\sumnodeD_{j}\cj} = \sum_{j}\E{\sumnodeD_{j}}\cj \, . \label{eq:exp-y}
\end{align}
For the variances we obtain:
\begin{align}
  \Var{A} &= \Var{\sumnodeD_i}\ci^{2} \label{eq:var-x}\\
  \Var{B} &= \sum_{j}\Var{\sumnodeD_j}\cj^{2} + \sum_{j_1 \neq j_2} \Cov{\sumnodeD_{j_{1}}, \sumnodeD_{j_{2}}}c_{j_{1}}c_{j_{2}} \, . \label{eq:var-y}
\end{align}
Following \cref{eq:sum-covariance}, the covariance term between a root node and the sum of all root nodes can be decomposed as follows:
\begin{align}
  \label{eq:cov-x-y}
  \Cov{A, B} &= \ci \sum_{j} \cj \Cov{\sumnodeD_{i},\sumnodeD_{j}} \quad ,
\end{align}
which in turn can be resolved with one of the methods provided in \cref{sec:tdi:derivation:covariance}.

While \cref{eq:exp-rv-ratio,eq:var-rv-ratio} are seemingly simple, their particular
formulation implies statistical independence between $A$ and $B$. Since $B$ is
a sum over all $A$, this independence naturally does not hold. Therefore, the
solution given here is only an approximation of the true second-order Taylor
approximation. In Appendix A.2.7 we extend the formulation of
\cite{ratio_seltman} and take into account the dependencies between root nodes $\sumnode_i$ and their sum $\sum_{i}\sumnode_i$. 

\subsubsection{Tractability}
\label{sec:tdi:derivation:tractability}
We re-emphasize that PCs are tractable probabilistic models where, in general,
inference is at most polynomial in the network size.
Specifically, in the PC family, SPNs perform a wide range of queries in linear time
in the network size.

Thanks to the compact representation of PCs, all formulations derived for TDI in
\cref{sec:tdi:derivation:expectation,sec:tdi:derivation:variance,sec:tdi:derivation:covariance,sec:tdi:derivation:leaf-nodes,sec:tdi:derivation:classification}
have polynomial space and time complexity, specifically, at most quadratic.
In fact, the expectation of a single sum node, given by \cref{eq:sum-expectation}, adds a negligible single
floating point multiplication (by $q$) as we can reuse the sum node output. The expectation
of a single product, in \cref{eq:prod-expectation}, does not require any additional operations. For the
computation of variance and covariance, the cost depends on the actual PC structure.
In sparse structures such as trees, the cost is linear with respect to the number of
input nodes, as the covariance term becomes zero.
However, for cases where all child input covariance combinations need to be computed, as in \cref{eq:prod-variance}, the cost can be
locally quadratic with respect to the number of sum node inputs (see \cref{eq:sum-variance,eq:sum-covariance}). Applying solution c) of
\cref{sec:tdi:derivation:covariance}, i.e.,
the implicit ``copy-paste'' augmentation of the DAG, reduces the cost of \cref{eq:prod-variance}
to be linear instead of quadratic.
This renders a full bottom-up pass tractable,
which can be performed in parallel with the standard bottom-up
probabilistic inference procedure. We provide pseudocode for the bottom-up TDI in Appendix B.

\section{Experimental Evaluation}
\label{sec:experimental_evaluation}
To demonstrate that PCs generally suffer from overconfidence under various forms of
distribution shift, and to show the benefits of TDI in these circumstances,
we investigate the following three common experimental scenarios:

\begin{enumerate}
    \item \textbf{OOD datasets:} Following popular practice to assess whether a
    model can successfully distinguish known data from unknown OOD instances
    \citep{bradshaw_2017, nalisnick_2019}, we train the circuits on SVHN~\citep{svhn} and then additionally test on several popular color image datasets: CIFAR-100~\citep{cifar100}, CINIC~\citep{cinic}, and LSUN~\citep{lsun}.
    \item \textbf{Perturbations:} Inspired by recent works that investigate predictions in the context of increasingly perturbed data instances \citep{ovadia_2019, antoran_2020, daxberger_2021}, we evaluate our models when rotating MNIST \citep{lecun_1998} digits are introduced for inference.
    \item \textbf{Corrupted inputs:} In the spirit of recent works that
    demonstrate standard neural networks' inability to effectively handle
    naturally corrupted data \citep{hendrycks_2019, michaelis_2019}, we
    include a set of 15 different non-trivial corruptions to the SVHN dataset
    for inference with PCs and PCs + TDI. Each of these corruptions features five different levels of severity.
\end{enumerate}

\paragraph{Experimental Setup.}
For our experiments, we implemented TDI based on RAT-SPNs in PyTorch and \emph{SPFlow}~\citep{molina_2019}. We use $S=20, I=20, D=5, R=5$ for the RAT-SPN structure and train our models for 200 epochs with a mini-batch size of 200, a learning rate of 1e-3 with the Adam~\citep{kingma2015adam} optimizer, and a PC + TDI dropout value of 0.2 for MNIST and 0.1 for SVHN. 
A detailed description is provided in Appendix C and our code is available at \url{https://github.com/ml-research/tractable-dropout-inference}.

\subsection{PCs with TDI Detect OOD Data}
Following our outlined first scenario, we first train on the SVHN dataset. We then
evaluate the predictive entropy obtained on samples of the unseen test set and on instances that come from entirely different
distributions of other datasets, e.g. house numbers vs. different
scene images or object categories like cars and sofas.
To successfully avoid mispredictions on an unrelated unknown dataset,
the entropy of our model's predictions should be higher compared to the one obtained for
ID samples.
Although predictive entropy might not be the optimal choice
for uncertainty quantification \citep{hullermeier_2021}, we leave the exploration
of better measures, in particular for OOD detection, to future work and remark that
our main focus is on assessing whether the epistemic uncertainty estimate obtained with TDI
is meaningful or not.

\begin{table}
    \caption{PC + TDI (in bold) improves area under curve scores for \cref{fig:precision-ood} over PCs when measuring the OOD precision over all thresholds by more than 2$\times$ on every OOD dataset. TDI is competitive with MCD while being more efficient as it does not require the cost of performing multiple (here 100) stochastic forward passes.
    }
    \centering
    \begin{tabular}{r|rrr}
          AUC ($\uparrow$) & CIFAR & CINIC & LSUN \\
            \midrule
         PC & 29.3 & 29.9 & 30.3 \\
         PC + TDI & \textbf{64.6} & \textbf{66.1} & \textbf{81.8} \\
         PC + MCD & 68.5 & 70.0 & 84.9 \\
         \midrule
         MLP & 56.0 & 58.1 & 55.9 \\
         MLP + MCD & 80.8 & 95.9 & 80.5 \\
         LeNet & 17.1 & 15.6 & 2.6 \\
         LeNet + MCD & 43.4 & 92.9 & 29.5 \\
    \end{tabular}
    \label{tab:ood-auc-scores}
\end{table}

\begin{figure}
    \centering
    \input{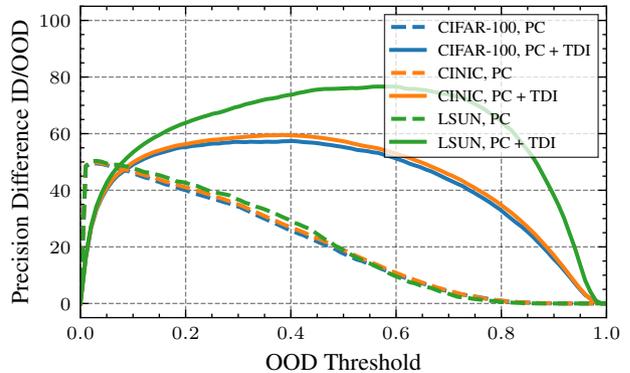}
    \caption{Difference between ID and OOD detection precision of PCs (dashed) and PCs + TDI (solid) for all OOD detection threshold of \cref{fig:precision-ood}. PCs with TDI both outperform PCs in absolute terms and allow the model to adequately balance ID versus OOD data. On the contrary, conventional PCs without TDI generally perform poorly and have their best point at an unintuitively low threshold.}
    \label{fig:precision-ood-diffs}
\end{figure}

Our introductory \cref{fig:precision-ood} has already shown that conventional PCs are bad at properly detecting OOD data while keeping a high precision on ID data, whereas PCs with TDI overcome this challenge. To also quantify the improvement introduced by TDI over all thresholds, we now show the area under the curve scores for \cref{fig:precision-ood} in \cref{tab:ood-auc-scores}, demonstrating that TDI improves all scenarios by more than two times.
For a comprehensive overview and as a point of reference, we also present results illustrating that MCD attains analogous improvements on both PCs and neural networks such as a Multilayer Perceptron (MLP) and an inherently overconfident LeNet. However, while MCD requires several stochastic forward passes (100 in this case), by exploiting PCs semantics, TDI achieves similar performance with only a single forward pass. 

In addition, we further highlight the precise tradeoff between the ID and OOD precision over all OOD decision thresholds in \cref{fig:precision-ood-diffs}. In other words, we quantify the precision with which a selected threshold on entropy correctly leads to rejection of unknown OOD data, while at the same time \emph{not} rejecting ID data in order to classify it correctly. Intuitively, the threshold should balance the latter two, as a very low threshold should simply reject all data, whereas a very high threshold would incorrectly accept any inputs. As visible, this is not the case for PCs, that have their largest margin between ID and OOD error at a very low OOD decision threshold, leading to a high ID error, e.g. 28.7\% ID error and 21.8\% LSUN OOD error at a threshold of 0.05 (cf. \cref{fig:precision-ood}). On the contrary, TDI balances this shortcoming and allows for much higher OOD decision thresholds while keeping a lower ID error, e.g. 13.2\% ID error and 10.2\% LSUN OOD error at a reasonable mid-way threshold of 0.6 (cf. \cref{fig:precision-ood}). A complementary view with predictive entropy and uncertainty in terms of the standard deviation as the square root of \cref{eq:var-rv-ratio} is provided in Appendix D.

\subsection{PCs with TDI are More Uncertain on Perturbed Samples}
\begin{figure}[t]
  \centering
  \input{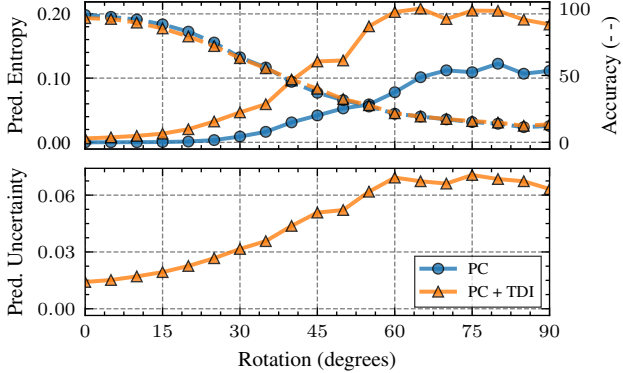}
  \caption{Top panel: predictive entropy (left y-axis, solid) and accuracy
    (right y-axis, dashed) of a PC (blue circles) and PC + TDI (orange
    triangles) on a gradual distribution shift of increasing MNIST digit
    rotations from 0$^\circ$ to 90$^\circ$ (x-axis). TDI already captures the
    distribution shift at lower degrees of rotation and assigns a much larger
    predictive entropy to greater rotations than PCs, while retaining
    predictive accuracy. Bottom panel: complementary view of predictive
    uncertainty (standard deviation in \cref{eq:var-rv-ratio}) in PC + TDI.}
    \label{fig:rotating_mnist_digits}
\end{figure}
\begin{figure*}[t]
  \centering
  \input{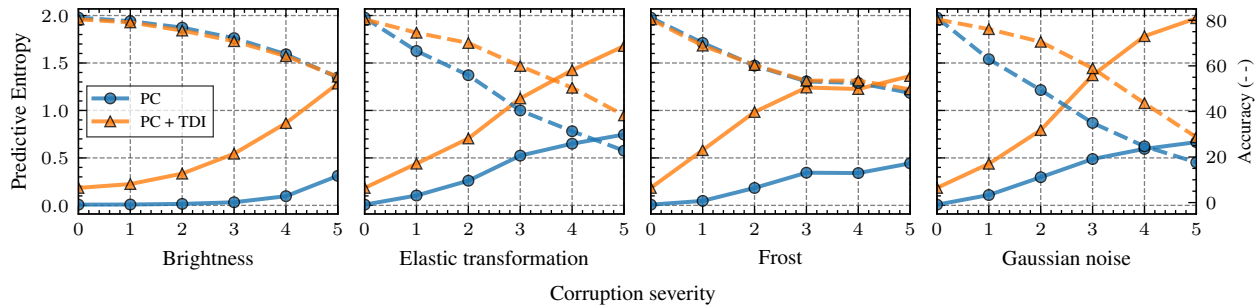}
  \caption{
  Predictive entropy (left y-axis) and accuracy (right y-axis) of a PC (blue
   circles) and PC + TDI (orange triangles) for increasingly corrupted SVHN data at five severity levels; here, altering brightness, introducing elastic transformation, simulating frost, and adding Gaussian noise. PCs with TDI detect the distribution shift by assigning higher predictive entropy with increasing severity, while at the same time being more robust in predictive accuracy against the corruption.}
    \label{fig:svhn_corruptions}
\end{figure*}
In addition to the abrupt distribution shift of the prior section, we now
inspect the behavior of PCs and PCs + TDI on a more gradual scale in the second
scenario. Here, we train on the original MNIST training set. At inference time, we evaluate the models' predictive entropy on rotated versions of the MNIST test set from 0$^\circ$ to 90$^\circ$ in steps of 5$^\circ$, to simulate a gradual increase in data perturbation. Once more, with the aim of an effective measure for the distribution shift, the model should assign higher predictive entropy and uncertainty with increasing data perturbation. We visualize this experiment in \cref{fig:rotating_mnist_digits}, demonstrating in the top panel, that PCs with TDI (orange triangle) measure the data perturbation already at lower degrees of rotation and assigns a higher entropy to larger rotations than PCs (blue, circles). At the same time, TDI retains the same predictive accuracy. In the bottom panel of \cref{fig:rotating_mnist_digits}, we additionally highlight the measure of uncertainty as the standard deviation of \cref{eq:var-rv-ratio}, again confirming the expected increase of uncertainty with increasing data perturbation from an auxiliary viewpoint.

\subsection{PCs with TDI are More Robust to Data Corruptions}
As outlined in our third scenario, we investigate the case of natural and
synthetic data corruptions. We train on the SVHN training set and then evaluate the models on corrupted versions of the SVHN test set, with 15 different corruptions at five increasing levels of severity. Similar to the prior two scenarios, a successful detection entails that the models should be able to attribute a progressive increase in predictive entropy with increasing corruption severity. In \cref{fig:svhn_corruptions} we highlight the model's behavior on four such corruption types: brightness, elastic transformation, simulated frost, and Gaussian noise (see Appendix E for analogous evidence for all 15 corruption types). Matching the behavior of the perturbation scenario, in all shown corruption settings, a PC with TDI can associate an increase in corruption severity with higher predictive entropy. On top of that, TDI stays more robust at all severity levels of corruption than the PC by retaining higher predictive accuracy in the case of elastic transformation and Gaussian noise corruptions. This third scenario thus further verifies TDI's robustness against, and their ability to capture distribution shift.

\subsection{Discussion}
Our empirical evidence across all three scenarios indicates that PC + TDI is in
fact more robust and provides model uncertainty estimates that allow detecting
data from various unknown and shifted distributions. TDI lets PCs \emph{``know what they don't know''}. Beyond this desideratum, the sampling-free uncertainty of TDI entails several advantages over MCD in neural networks, opening up various additional prospects.

\textbf{Prospects:}
On the one hand, TDI alleviates the computational burden of MCD, getting rid of the compromise between estimation quality and amount of forward passes.
This can be particularly beneficial to other circuits such as regression or logistic circuits~\citep{liang2019logistic-circuits} tailored to predictive tasks.
Moreover, the tractable computation in a single forward pass in turn paves the way for uncertainty estimates to be directly involved in training processes. Such a signal does not only help with robustness but can also improve
active or continual learning,
in which PCs are largely yet to be explored. On the other hand, the clear semantics of PCs allow for the prospective inclusion of prior knowledge about uncertainty through the explicit (co-)variance terms at leaf nodes (recall \cref{sec:tdi:derivation:leaf-nodes}), which is unavailable to neural networks.

\textbf{Limitations:}
Whereas TDI has removed the computational burden of MCD, the necessity to select a dropout chance $p$, as a hyperparameter, remains untouched. In our various experiments, and prior works in neural networks, a common low value seems to suffice, but it is an additional consideration to be taken into account for training and inference. On the empirical side, further investigation of TDI should be extended to arbitrary structures,
involving the propagation of the covariance as introduced in~\cref{sec:tdi:derivation:covariance}. In similar spirit, although we have already experimentally investigated three distinct scenarios, the empirical performance of TDI for other density estimation tasks remains to be explored. 

\section{Conclusion}
In the spirit of recent works for neural networks, we have highlighted that the
generative model family of PCs suffers from overconfidence and is thus unable to
effectively separate ID from OOD data. As a remedy to this challenge, we have
drawn inspiration from the well-known MCD and introduced a novel probabilistic
inference method capable of providing tractable uncertainty estimates: tractable
dropout inference.
We obtain such sampling-free, single computation pass estimates by deriving a closed-form solution through variance propagation. Our empirical evidence confirms that TDI provides improved robustness and comes with the ability to detect distribution changes in three key scenarios: dataset change, data perturbation, and data corruption. The computationally cheap nature and potential to include prior knowledge in TDI paves the way for various future work, such as including uncertainty in training.

\begin{acknowledgements} %
This work was supported by the Federal Ministry of Education and Research (BMBF) Competence Center for AI and Labour (``kompAKI'', FKZ 02L19C150) and the project ``safeFBDC - Financial Big
Data Cluster'' (FKZ: 01MK21002K), funded by the German Federal Ministry for
Economics Affairs and Energy as part of the GAIA-x initiative. It benefited from the Hessian
Ministry of Higher Education, Research, Science and the Arts (HMWK; projects
``The Third Wave of AI'' and ``The Adaptive Mind''), and the Hessian research
priority programme LOEWE within the project ``WhiteBox''.
\end{acknowledgements}

\bibliography{bibliography.bib}

\begin{thebibliography}{61}
\providecommand{\natexlab}[1]{#1}
\providecommand{\url}[1]{\texttt{#1}}
\expandafter\ifx\csname urlstyle\endcsname\relax
  \providecommand{\doi}[1]{doi: #1}\else
  \providecommand{\doi}{doi: \begingroup \urlstyle{rm}\Url}\fi

\bibitem[Adel et~al.(2015)Adel, Balduzzi, and Ghodsi]{adel_2015}
Tameem Adel, David Balduzzi, and Ali Ghodsi.
\newblock Learning the structure of sum-product networks via an svd-based
  algorithm.
\newblock In \emph{UAI}, 2015.

\bibitem[Amodei et~al.(2016)Amodei, Olah, Steinhardt, Christiano, Schulman, and
  Man{\'{e}}]{amodei_2016}
Dario Amodei, Chris Olah, Jacob Steinhardt, Paul~F. Christiano, John Schulman,
  and Dan Man{\'{e}}.
\newblock Concrete problems in {AI} safety.
\newblock \emph{arXiv preprint arXiv:1606.06565}, 2016.

\bibitem[Antor{\'{a}}n et~al.(2020)Antor{\'{a}}n, Allingham, and
  Hern{\'{a}}ndez{-}Lobato]{antoran_2020}
Javier Antor{\'{a}}n, James~Urquhart Allingham, and Jos{\'{e}}~Miguel
  Hern{\'{a}}ndez{-}Lobato.
\newblock Depth uncertainty in neural networks.
\newblock In \emph{NeurIPS}, 2020.

\bibitem[Blundell et~al.(2015)Blundell, Cornebise, Kavukcuoglu, and
  Wierstra]{blundell_2015}
Charles Blundell, Julien Cornebise, Koray Kavukcuoglu, and Daan Wierstra.
\newblock Weight uncertainty in neural network.
\newblock In \emph{ICML}, 2015.

\bibitem[Boult et~al.(2019)Boult, Cruz, Dhamija, Gunther, Henrydoss, and
  Scheirer]{boult_2019}
Terrance~E. Boult, Steve Cruz, Akshay~Raj Dhamija, Manuel Gunther, James
  Henrydoss, and Walter~J. Scheirer.
\newblock Learning and the unknown: Surveying steps toward open world
  recognition.
\newblock In \emph{AAAI}, 2019.

\bibitem[Bradshaw et~al.(2017)Bradshaw, Matthews, and
  Ghahramani]{bradshaw_2017}
John Bradshaw, Alexander G de~G Matthews, and Zoubin Ghahramani.
\newblock Adversarial examples, uncertainty, and transfer testing robustness in
  gaussian process hybrid deep networks.
\newblock \emph{arXiv preprint arXiv:1707.02476}, 2017.

\bibitem[Cerutti et~al.(2022)Cerutti, Kaplan, Kimmig, and Sensoy]{cerutti_2022}
Federico Cerutti, Lance~M. Kaplan, Angelika Kimmig, and Murat Sensoy.
\newblock Handling epistemic and aleatory uncertainties in probabilistic
  circuits.
\newblock \emph{Machine Learning}, 2022.

\bibitem[Choi et~al.(2020)Choi, Vergari, and Van~den Broeck]{choi_2020}
YooJung Choi, Antonio Vergari, and Guy Van~den Broeck.
\newblock Probabilistic circuits: A unifying framework for tractable
  probabilistic models.
\newblock Technical report, UCLA, 2020.

\bibitem[Darlow et~al.(2018)Darlow, Crowley, Antoniou, and Storkey]{cinic}
Luke~Nicholas Darlow, Elliot~J. Crowley, Antreas Antoniou, and Amos~J. Storkey.
\newblock {CINIC-10} is not imagenet or {CIFAR-10}.
\newblock \emph{arXiv preprint arXiv:1810.03505}, 2018.

\bibitem[Daxberger et~al.(2021)Daxberger, Nalisnick, Allingham, Antor{\'{a}}n,
  and Hern{\'{a}}ndez{-}Lobato]{daxberger_2021}
Erik~A. Daxberger, Eric~T. Nalisnick, James~Urquhart Allingham, Javier
  Antor{\'{a}}n, and Jos{\'{e}}~Miguel Hern{\'{a}}ndez{-}Lobato.
\newblock Bayesian deep learning via subnetwork inference.
\newblock In \emph{ICML}, 2021.

\bibitem[Delalleau and Bengio(2011)]{delalleau_2011}
Olivier Delalleau and Yoshua Bengio.
\newblock Shallow vs. deep sum-product networks.
\newblock In \emph{NIPS}, 2011.

\bibitem[{Deratani Mauá} et~al.(2018){Deratani Mauá}, Conaty, {Gagliardi
  Cozman}, Poppenhaeger, and {Polpo de Campos}]{maua_2017}
Denis {Deratani Mauá}, Diarmaid Conaty, Fabio {Gagliardi Cozman}, Katja
  Poppenhaeger, and Cassio {Polpo de Campos}.
\newblock Robustifying sum-product networks.
\newblock \emph{International Journal of Approximate Reasoning}, 2018.

\bibitem[Ebrahimi et~al.(2020)Ebrahimi, Elhoseiny, Darrell, and
  Rohrbach]{ebrahimi_2020}
Sayna Ebrahimi, Mohamed Elhoseiny, Trevor Darrell, and Marcus Rohrbach.
\newblock Uncertainty-guided continual learning with bayesian neural networks.
\newblock In \emph{ICLR}, 2020.

\bibitem[Gal and Ghahramani(2016)]{gal_2016}
Yarin Gal and Zoubin Ghahramani.
\newblock Dropout as a bayesian approximation: Representing model uncertainty
  in deep learning.
\newblock In \emph{ICML}, 2016.

\bibitem[Gal et~al.(2017)Gal, Islam, and Ghahramani]{gal_2017}
Yarin Gal, Riashat Islam, and Zoubin Ghahramani.
\newblock Deep bayesian active learning with image data.
\newblock In \emph{ICML}, 2017.

\bibitem[Gens and Domingos(2013)]{gens_2013}
Robert Gens and Pedro Domingos.
\newblock {Learning the Structure of Sum-Product Networks}.
\newblock In \emph{ICML}, 2013.

\bibitem[Graves(2011)]{graves_2011}
Alex Graves.
\newblock Practical variational inference for neural networks.
\newblock In \emph{NIPS}, 2011.

\bibitem[Guo et~al.(2017)Guo, Pleiss, Sun, and Weinberger]{guo_2017}
Chuan Guo, Geoff Pleiss, Yu~Sun, and Kilian~Q. Weinberger.
\newblock On calibration of modern neural networks.
\newblock In \emph{ICML}, 2017.

\bibitem[Hansen and Salamon(1990)]{hansen_1990}
Lars~Kai Hansen and Peter Salamon.
\newblock Neural network ensembles.
\newblock \emph{TPAMI}, 1990.

\bibitem[Hendrycks and Dietterich(2019)]{hendrycks_2019}
Dan Hendrycks and Thomas~G. Dietterich.
\newblock Benchmarking neural network robustness to common corruptions and
  perturbations.
\newblock In \emph{ICLR}, 2019.

\bibitem[Hern{\'{a}}ndez{-}Lobato and Adams(2015)]{hernandez_lobato_2015}
Jos{\'{e}}~Miguel Hern{\'{a}}ndez{-}Lobato and Ryan~P. Adams.
\newblock Probabilistic backpropagation for scalable learning of bayesian
  neural networks.
\newblock In \emph{ICML}, 2015.

\bibitem[H{\"{u}}llermeier and Waegeman(2021)]{hullermeier_2021}
Eyke H{\"{u}}llermeier and Willem Waegeman.
\newblock Aleatoric and epistemic uncertainty in machine learning: an
  introduction to concepts and methods.
\newblock \emph{Machine Learning}, 2021.

\bibitem[Izmailov et~al.(2021)Izmailov, Vikram, Hoffman, and
  Wilson]{izmailov_2021}
Pavel Izmailov, Sharad Vikram, Matthew~D Hoffman, and Andrew Gordon~Gordon
  Wilson.
\newblock What are bayesian neural network posteriors really like?
\newblock In \emph{ICML}, 2021.

\bibitem[Kendall and Gal(2017)]{kendall_2017}
Alex Kendall and Yarin Gal.
\newblock What uncertainties do we need in bayesian deep learning for computer
  vision?
\newblock In \emph{NeurIPS}, 2017.

\bibitem[Kendall et~al.(2017)Kendall, Badrinarayanan, and
  Cipolla]{kendall2017a}
Alex Kendall, Vijay Badrinarayanan, and Roberto Cipolla.
\newblock Bayesian segnet: Model uncertainty in deep convolutional
  encoder-decoder architectures for scene understanding.
\newblock In \emph{BMVC}, 2017.

\bibitem[Khan et~al.(2019)Khan, Immer, Abedi, and Korzepa]{khan_2019}
Mohammad~Emtiyaz Khan, Alexander Immer, Ehsan Abedi, and Maciej Korzepa.
\newblock Approximate inference turns deep networks into gaussian processes.
\newblock In \emph{NeurIPS}, 2019.

\bibitem[Kingma and Ba(2015)]{kingma2015adam}
Diederik~P. Kingma and Jimmy Ba.
\newblock Adam: A method for stochastic optimization.
\newblock In \emph{ICLR}, 2015.

\bibitem[Kingma and Welling(2014)]{kingma2014auto}
Diederik~P. Kingma and Max Welling.
\newblock Auto-encoding variational bayes.
\newblock In \emph{ICLR}, 2014.

\bibitem[Kobyzev et~al.(2020)Kobyzev, Prince, and
  Brubaker]{kobyzev2020normalizing}
Ivan Kobyzev, Simon~JD Prince, and Marcus~A Brubaker.
\newblock Normalizing flows: An introduction and review of current methods.
\newblock \emph{TPAMI}, 2020.

\bibitem[Krizhevsky(2009)]{cifar100}
Alex Krizhevsky.
\newblock Learning multiple layers of features from tiny images.
\newblock Technical report, U. of Toronto, 2009.

\bibitem[Lakshminarayanan et~al.(2017)Lakshminarayanan, Pritzel, and
  Blundell]{lakshminarayanan_2017}
Balaji Lakshminarayanan, Alexander Pritzel, and Charles Blundell.
\newblock Simple and scalable predictive uncertainty estimation using deep
  ensembles.
\newblock In \emph{NeurIPS}, 2017.

\bibitem[LeCun et~al.(1998)LeCun, Bottou, Bengio, and Haffner]{lecun_1998}
Yann LeCun, L{\'{e}}on Bottou, Yoshua Bengio, and Patrick Haffner.
\newblock Gradient-based learning applied to document recognition.
\newblock \emph{Proc. {IEEE}}, 1998.

\bibitem[Liang and {Van den Broeck}(2019)]{liang2019logistic-circuits}
Yitao Liang and Guy {Van den Broeck}.
\newblock Learning logistic circuits.
\newblock In \emph{AAAI}, 2019.

\bibitem[Louizos and Welling(2016)]{louizos_2016}
Christos Louizos and Max Welling.
\newblock Structured and efficient variational deep learning with matrix
  gaussian posteriors.
\newblock In \emph{ICML}, 2016.

\bibitem[MacKay(1992)]{mackay_1992}
David~JC MacKay.
\newblock A practical bayesian framework for backpropagation networks.
\newblock \emph{Neural computation}, 1992.

\bibitem[Maddox et~al.(2019)Maddox, Izmailov, Garipov, Vetrov, and
  Wilson]{maddox_2019}
Wesley~J. Maddox, Pavel Izmailov, Timur Garipov, Dmitry~P. Vetrov, and
  Andrew~Gordon Wilson.
\newblock A simple baseline for bayesian uncertainty in deep learning.
\newblock In \emph{NeurIPS}, 2019.

\bibitem[Matan et~al.(1990)Matan, Kiang, Stenard, Boser, Denker, Henderson,
  Howard, Hubbard, Jackel, and Le~Cun]{matan_90}
Ofer Matan, RK~Kiang, CE~Stenard, B~Boser, JS~Denker, Don Henderson, RE~Howard,
  W~Hubbard, LD~Jackel, and Yann Le~Cun.
\newblock Handwritten character recognition using neural network architectures.
\newblock In \emph{USPS advanced technology conference}, 1990.

\bibitem[Michaelis et~al.(2019)Michaelis, Mitzkus, Geirhos, Rusak, Bringmann,
  Ecker, Bethge, and Brendel]{michaelis_2019}
Claudio Michaelis, Benjamin Mitzkus, Robert Geirhos, Evgenia Rusak, Oliver
  Bringmann, Alexander~S. Ecker, Matthias Bethge, and Wieland Brendel.
\newblock Benchmarking robustness in object detection: Autonomous driving when
  winter is coming.
\newblock \emph{arXiv preprint arXiv:1907.07484}, 2019.

\bibitem[Miller et~al.(2018)Miller, Nicholson, Dayoub, and
  S{\"u}nderhauf]{miller_2018}
Dimity Miller, Lachlan Nicholson, Feras Dayoub, and Niko S{\"u}nderhauf.
\newblock Dropout sampling for robust object detection in open-set conditions.
\newblock In \emph{ICRA}, 2018.

\bibitem[Mishkin et~al.(2018)Mishkin, Kunstner, Nielsen, Schmidt, and
  Khan]{mishkin_2018}
Aaron Mishkin, Frederik Kunstner, Didrik Nielsen, Mark Schmidt, and
  Mohammad~Emtiyaz Khan.
\newblock {SLANG:} fast structured covariance approximations for bayesian deep
  learning with natural gradient.
\newblock In \emph{NeurIPS}, 2018.

\bibitem[Molina et~al.(2019)Molina, Vergari, Stelzner, Peharz, Subramani,
  Mauro, Poupart, and Kersting]{molina_2019}
Alejandro Molina, Antonio Vergari, Karl Stelzner, Robert Peharz, Pranav
  Subramani, Nicola~Di Mauro, Pascal Poupart, and Kristian Kersting.
\newblock Spflow: An easy and extensible library for deep probabilistic
  learning using sum-product networks.
\newblock \emph{arXiv preprint arXiv:1901.03704}, 2019.

\bibitem[Mundt et~al.(2022)Mundt, Pliushch, Majumder, Hong, and
  Ramesh]{mundt_2022}
Martin Mundt, Iuliia Pliushch, Sagnik Majumder, Yongwon Hong, and Visvanathan
  Ramesh.
\newblock Unified probabilistic deep continual learning through generative
  replay and open set recognition.
\newblock \emph{Journal of Imaging}, 2022.

\bibitem[Nalisnick et~al.(2019)Nalisnick, Matsukawa, Teh, G{\"{o}}r{\"{u}}r,
  and Lakshminarayanan]{nalisnick_2019}
Eric~T. Nalisnick, Akihiro Matsukawa, Yee~Whye Teh, Dilan G{\"{o}}r{\"{u}}r,
  and Balaji Lakshminarayanan.
\newblock Do deep generative models know what they don't know?
\newblock In \emph{ICLR}, 2019.

\bibitem[Neal(2012)]{neal_2012}
Radford~M Neal.
\newblock \emph{Bayesian learning for neural networks}.
\newblock Springer Science \& Business Media, 2012.

\bibitem[Netzer et~al.(2011)Netzer, Wang, Coates, Bissacco, Wu, and Ng]{svhn}
Yuval Netzer, Tao Wang, Adam Coates, Alessandro Bissacco, Bo~Wu, and Andrew~Y.
  Ng.
\newblock Reading digits in natural images with unsupervised feature learning.
\newblock In \emph{NIPS Workshop on Deep Learning and Unsupervised Feature
  Learning}, 2011.

\bibitem[Nguyen et~al.(2015)Nguyen, Yosinski, and Clune]{nguyen_2015}
Anh~Mai Nguyen, Jason Yosinski, and Jeff Clune.
\newblock Deep neural networks are easily fooled: High confidence predictions
  for unrecognizable images.
\newblock In \emph{CVPR}, 2015.

\bibitem[Osband et~al.(2021)Osband, Wen, Asghari, Ibrahimi, Lu, and
  Roy]{osband_2021}
Ian Osband, Zheng Wen, Mohammad Asghari, Morteza Ibrahimi, Xiyuan Lu, and
  Benjamin~Van Roy.
\newblock Epistemic neural networks.
\newblock \emph{arXiv preprint arXiv:2107.08924}, 2021.

\bibitem[Ovadia et~al.(2019)Ovadia, Fertig, Ren, Nado, Sculley, Nowozin,
  Dillon, Lakshminarayanan, and Snoek]{ovadia_2019}
Yaniv Ovadia, Emily Fertig, Jie Ren, Zachary Nado, David Sculley, Sebastian
  Nowozin, Joshua Dillon, Balaji Lakshminarayanan, and Jasper Snoek.
\newblock Can you trust your model's uncertainty? evaluating predictive
  uncertainty under dataset shift.
\newblock In \emph{NeurIPS}, 2019.

\bibitem[Papamakarios et~al.(2021)Papamakarios, Nalisnick, Rezende, Mohamed,
  and Lakshminarayanan]{papamakarios2021normalizing}
George Papamakarios, Eric Nalisnick, Danilo~Jimenez Rezende, Shakir Mohamed,
  and Balaji Lakshminarayanan.
\newblock Normalizing flows for probabilistic modeling and inference.
\newblock \emph{JMLR}, 2021.

\bibitem[Peddi et~al.(2022)Peddi, Rahman, and Gogate]{peddi_2022}
Rohith Peddi, Tahrima Rahman, and Vibhav Gogate.
\newblock Robust learning of tractable probabilistic models.
\newblock In \emph{UAI}, 2022.

\bibitem[Peharz et~al.(2015)Peharz, Tschiatschek, Pernkopf, and
  Domingos]{peharz_2015}
Robert Peharz, Sebastian Tschiatschek, Franz Pernkopf, and Pedro Domingos.
\newblock On theoretical properties of sum-product networks.
\newblock In \emph{AISTATS}, 2015.

\bibitem[Peharz et~al.(2020{\natexlab{a}})Peharz, Lang, Vergari, Stelzner,
  Molina, Trapp, den Broeck, Kersting, and Ghahramani]{peharz_2020_icml}
Robert Peharz, Steven Lang, Antonio Vergari, Karl Stelzner, Alejandro Molina,
  Martin Trapp, Guy~Van den Broeck, Kristian Kersting, and Zoubin Ghahramani.
\newblock Einsum networks: Fast and scalable learning of tractable
  probabilistic circuits.
\newblock In \emph{ICML}, 2020{\natexlab{a}}.

\bibitem[Peharz et~al.(2020{\natexlab{b}})Peharz, Vergari, Stelzner, Molina,
  Shao, Trapp, Kersting, and Ghahramani]{peharz_2020}
Robert Peharz, Antonio Vergari, Karl Stelzner, Alejandro Molina, Xiaoting Shao,
  Martin Trapp, Kristian Kersting, and Zoubin Ghahramani.
\newblock Random sum-product networks: A simple and effective approach to
  probabilistic deep learning.
\newblock In \emph{UAI}, 2020{\natexlab{b}}.

\bibitem[Poon and Domingos(2011)]{poon_2011}
Hoifung Poon and Pedro~M. Domingos.
\newblock Sum-product networks: {A} new deep architecture.
\newblock In \emph{UAI}, 2011.

\bibitem[Rooshenas and Lowd(2014)]{rooshenas_2014}
Amirmohammad Rooshenas and Daniel Lowd.
\newblock Learning sum-product networks with direct and indirect variable
  interactions.
\newblock In \emph{ICML}, 2014.

\bibitem[Scheirer et~al.(2013)Scheirer, de~Rezende~Rocha, Sapkota, and
  Boult]{scheirer_2013}
Walter~J Scheirer, Anderson de~Rezende~Rocha, Archana Sapkota, and Terrance~E
  Boult.
\newblock Toward open set recognition.
\newblock \emph{TPAMI}, 2013.

\bibitem[Scheirer et~al.(2014)Scheirer, Jain, and Boult]{scheirer_2014}
Walter~J. Scheirer, Lalit~P. Jain, and Terrance~E. Boult.
\newblock Probability models for open set recognition.
\newblock \emph{TPAMI}, 2014.

\bibitem[Seltman(2018)]{ratio_seltman}
Howard Seltman.
\newblock Approximations for mean and variance of a ratio.
\newblock Technical report, CMU, 2018.

\bibitem[Srivastava et~al.(2014)Srivastava, Hinton, Krizhevsky, Sutskever, and
  Salakhutdinov]{srivastava_2014}
Nitish Srivastava, Geoffrey Hinton, Alex Krizhevsky, Ilya Sutskever, and Ruslan
  Salakhutdinov.
\newblock Dropout: A simple way to prevent neural networks from overfitting.
\newblock \emph{JMLR}, 2014.

\bibitem[Yu et~al.(2015)Yu, Zhang, Song, Seff, and Xiao]{lsun}
Fisher Yu, Yinda Zhang, Shuran Song, Ari Seff, and Jianxiong Xiao.
\newblock {LSUN:} construction of a large-scale image dataset using deep
  learning with humans in the loop.
\newblock \emph{arXiv preprint arXiv:1506.03365}, 2015.

\bibitem[Yu et~al.(2022)Yu, Ventola, Thoma, Dhami, Mundt, and
  Kersting]{yu_2022_uai}
Zhongjie Yu, Fabrizio Ventola, Nils Thoma, Devendra~Singh Dhami, Martin Mundt,
  and Kristian Kersting.
\newblock Predictive whittle networks for time series.
\newblock In \emph{UAI}, 2022.

\end{thebibliography}

\ifarxiv
  \onecolumn %
  \section*{Supplementary Material}
  \appendix
  
Our paper's supplementary material contains various supporting materials and
complementary empirical evidence for our main paper’s findings. Specifically,
the appendix consists of six sections:

\begin{itemize}
  \item[\textbf{A}] \textbf{Tractable dropout inference derivations:} We provide
        the full derivations behind the presented TDI equations of our main
        body's Section 3.
  \item[\textbf{B}] \textbf{Tractable dropout inference pseudocode algorithm:}
        We present the pseudocode algorithm of the TDI procedure.
  \item[\textbf{C}] \textbf{Experimental setup:} The section contains additional
        details with respect to the experimental setup for our empirical
        evaluations of the main body.
  \item[\textbf{D}] \textbf{PCs with TDI can detect OOD data:} We provide a
        complementary view on the ability of TDI in detecting OOD data, showing
        the predictive entropy and the predictive uncertainty of a PC with TDI.
  \item[\textbf{E}] \textbf{PCs with TDI are more robust to corruptions:} We
        present additional quantitative evidence in form of the remaining
        corruptions not shown in the main body to further support our findings
        that PCs with TDI are more robust on corrupted data.
  \item[\textbf{F}] \textbf{Societal impact:} We briefly discuss the societal
        impact of our contributions.
\end{itemize}
\section{Tractable Dropout Inference Derivations}
\label{supp:appendix_a}
Appendix A contains the full set of derivations behind our closed-form solutions
of TDI. Accordingly, the subsequent subsections follow the notation and
structure of our main body's Section 3.

\subsection{Monte Carlo Dropout Uncertainty}
\label{sec:mc-dropout}
With Monte Carlo dropout we can estimate model uncertainty by measuring the
variance of the probability computed by the PC's root node. This is done by
replacing all sum node computations with:
\begin{align}
  \label{eq:sumnode-dropout}
  \sumnodeD &= \sum_{i}\delta_{i}w_{i}\ndi \, ,
\end{align}
where $\delta_{i} \sim \text{Bern}\!\left(1-p\right)$ is the result of a
Bernoulli trial with probability $1-p$. Here, $p$ is the dropout chance and
$\ndi$ is the value of the $i$-th child. Note that the product node computations
remain unchanged since they have no model parameters associated and simply
compute factorizations.
In the traditional Monte Carlo dropout applied to PCs, the forward pass for a
single input is then repeated $L$ times to finally compute the expected value
and variance of the root node (i.e., the PC probability):
\begin{align}
  \E{\nodeD_{\text{root}}} &= \frac{1}{L}\sum_{i=1}^{L} \nodeD_{\text{root},i} \, \\
  \Var{\nodeD_{\text{root}}} &= \frac{1}{L}\sum_{i=1}^{L} \left(\nodeD_{\text{root},i} - \E{\nodeD_{\text{root}}}\right)^{2} \, ,
\end{align}
where $\nodeD_{\text{root},i}$ is the $i$-th random Monte Carlo trial of
\cref{eq:sumnode-dropout}.

Since the Monte Carlo dropout formulation is based on repeated Bernoulli trials,
we can now look at it from a different perspective. We can define $\sumnode$ to
be a function of the Bernoulli random variable $\delta_{i}$. As the expectation
and variance for Bernoulli random variables are well known and easy to compute,
we only need to formulate how they propagate bottom-up through the PC in a
hierarchical manner, through sum and product nodes. We formally derive this in
the next section.

\subsection{TDI's Analytical Solution to Dropout Uncertainty}
\label{sec:cf-dropout}

To derive the analytical solution of expectation, variance, and covariance
propagation for TDI, we make use of the basic rules of how the expectation,
variance, and covariance behave under addition, multiplication, and constant
scaling. Recall that the goal of the derivation is to express the expectation,
variance, and covariance of a dropout node $\ndi$ in terms of the expectation,
variance, and covariance of its children
$\E{\ndj}, \Var{\ndj}, \Cov{\ndj, \nodeD_{j'}}$ for
$\ndj \in \text{children}\!\left( \ndi \right)$. This means that if we have a
closed-form solution for the expectation, variance, and covariance of the leaf
nodes, we can calculate the expectation, variance, and covariance of any
arbitrary node in the PC by the bottom-up propagation of $\E{\ndj}$,
$\Var{\ndj}$, and $\Cov{\ndj, \nodeD_{j'}}$.

\subsubsection{Expectation: The Point Estimate}
\label{sec:tdi:expectation}
\paragraph{Sum Nodes:}
Using the linearity of the expectation, we can move the expectation into the sum
and make use of the independence between the child nodes $\nodeD_{i}$ and the
Bernoulli RVs $\delta_{i}$ to extract $\Einline{\delta_i} = q$:
\begin{align}
  \E{\sumnodeD} &= \E{   \sum_{i} \delta_{i}w_{i}\ndi   } \\
                &=   \sum_{i} \E{\delta_{i}w_{i}\ndi} \\
                &=   \sum_{i} \E{\delta_{i}}w_{i}\E{\ndi} \\
                &=   \sum_{i} qw_{i}\E{\ndi} \\
                &=   q\sum_{i} w_{i}\E{\ndi} \label{eq:sumnode-final} \quad .
\end{align}
As of \cref{eq:sumnode-final}, we can see that the expected outcome of the sum
node is a convex combination with the original weights $w_{i}$ of the expected
outcome of its children $\ndi$, scaled by the expected drop in likelihood $q$.

\paragraph{Product Nodes:}
The decomposability of product nodes ensures the independence of their children
w.r.t. each other. This in turn leads to the product node expectation simply
becoming the product over the expectations of its children, i.e.,
\begin{align}
  \E{\prodnodeD} = \E{   \prod_{i} \ndi   } =    \prod_{i} \E{\ndi} \quad .
\end{align}

\subsubsection{Variance: The Uncertainty Proxy}
\label{sec:tdi:variance}
Similar to the original work of Monte Carlo dropout in neural networks
of~\cite{gal_2016}, we have used the variance as the proxy for the uncertainty.

\paragraph{Sum Nodes:}
The sum node variance decomposes into a sum of two terms. The first term is
based on the variances and expectations of its children and the second term
accounts for the covariance between the combinations of all children:
\begin{align}
  \label{eq:sum-node-variance}
  \Var{\sumnodeD} &= \Var{ \sum_{i} \delta_{i}w_{i}\ndi   } \\
                  &=    \sum_{i} \Var{ \delta_{i}w_{i}\ndi}  + \sum_{i\neq j}\Cov{\delta_{i}w_{i}\ndi, \delta_{j}w_{j}\ndj} \\
                  &=    \sum_{i} w_{i}^{2}\Var{\delta_{i}\ndi}  + \sum_{i\neq j}w_{i}w_{j}\Cov{\delta_{i}\ndi, \delta_{j}\ndj}\quad . \label{eq:sum-node-variance:last}
\end{align}
For the first term of \cref{eq:sum-node-variance:last}, we can further resolve
$\Var{ \delta_{i}\ndi }$ by employing the rules for the variance computation:
\begin{align}
  \Var{ \delta_{i}\ndi} &= \E{\delta_{i}^{2}}\E{\ndi^{2}} - \E{\delta_{i}}^{2}\E{\ndi}^{2}  \\
                        &= q\E{\ndi^{2}} - q^{2}\E{\ndi}^{2}  \\
                        &= q \left(  \Var{\ndi} + \E{\ndi}^{2}  \right) - q^{2}\E{\ndi}^{2}  \\
                        &= q \left( \Var{\ndi} + \E{\ndi}^{2}  - q\E{\ndi}^{2}  \right) \\
                        &= q \left( \Var{\ndi} + \left( 1 - q \right)\E{\ndi}^{2}  \right) \\
                        &= q \left( \Var{\ndi} + p\E{\ndi}^{2}  \right) \quad . \label{eq:sum-node-variance:left-subterm}
\end{align}
The second term of \cref{eq:sum-node-variance:last}, the summation over
$\Cov{\delta_{i}\ndi, \delta_{j}\ndj}$, includes the covariance between all
child nodes of $\sumnodeD$:
\begin{align}
  \label{eq:cov-sum-node-children}
  \Cov{\delta_{i}\ndi, \delta_{j}\ndj} &= \E{\delta_{i}\ndi \delta_{j}\ndj} - \E{\delta_{i}\ndi}\E{\delta_{j}\ndj} \\
                                       &= \E{\delta_{i}}\E{\delta_{j}}\E{\ndi\ndj} - \E{\delta_{i}}\E{\ndi}\E{\delta_{j}}\E{\ndj} \\
                                       &= q^{2} \left(\E{\ndi\ndj} - \E{\ndi}\E{\ndj} \right) \\
                                       &= q^{2} \Cov{\ndi, \ndj} \quad . \label{eq:cov-sum-node-children:last}
\end{align}
Thus, by putting the derivations of the two terms together, we obtain the
expression for the variance of a sum node:
\begin{align}
  \label{eq:sum-variance-deriv}
  \Var{\sumnodeD} = q\sum_{i} w_{i}^{2} \left( \Var{\nodeD_{i}} + p\E{\nodeD_{i}}^{2}  \right) + q^{2}\sum_{i\neq j}w_{i}w_{j}\Cov{\nodeD_{i}, \nodeD_{j}} \, .
\end{align}
\paragraph{Product Nodes:}
Similarly to the case of a sum node, the product node variance decomposes into
two product terms, by applying the product of independent variables rule, we
obtain:
\begin{align}
  \Var{\prodnodeD} &= \Var{ \prod_{i} \ndi} \\
                   &= \prod_{i} \E{\ndi^{2}} - \prod_{i} \E{\ndi}^{2} \\
                   &= \prod_{i} \left( \Var{\ndi} + \E{\ndi}^{2} \right) - \prod_{i} \E{\ndi}^{2} \quad .
\end{align}

\subsubsection{Covariance: The Evil}
\label{sec:tdi:covariance}
\paragraph{Sum Nodes:}
The covariance of two sum nodes, $\sumnodeD^{A}$ and $\sumnodeD^{B}$, neatly
decomposes into a weighted sum of all covariance combinations of the sum node
children, as shown in the following:
\begin{align}
  \label{eq:cov-two-sum-nodes}
  \Cov{\sumnodeD^{A}, \sumnodeD^{B}}  &= \Cov{\sum_{i}w_{i}^{A}\delta_{i}^{A}\nodeD_{i}^{A}, \sum_{j}w_{j}^{B}\delta_{j}^{B}\nodeD_{j}^{B}} \\
                                      &= \E{\sum_{i}w_{i}^{A}\delta_{i}^{A}\nodeD_{i}^{A} \sum_{j}w_{j}^{B}\delta_{j}^{B}\nodeD_{j}^{B}} - \E{\sum_{i}w_{i}^{A}\delta_{i}^{A}\nodeD_{i}^{A}}\E{\sum_{j}w_{j}^{B}\delta_{j}^{B}\nodeD_{j}^{B}} \\
                                      &= \sum_{i}w_{i}^{A}\sum_{j}w_{j}^{B}\E{\delta_{i}^{A}\nodeD_{i}^{A} \delta_{j}^{B}\nodeD_{j}^{B}} - \sum_{i}w_{i}^{A}\sum_{j}w_{j}^{B}\E{\delta_{i}^{A}\nodeD_{i}^{A}}\E{\delta_{j}^{B}\nodeD_{j}^{B}} \\
                                      &= \sum_{i}w_{i}^{A}\sum_{j}w_{j}^{B}\E{\delta_{i}^{A} \delta_{j}^{B}}\E{\nodeD_{i}^{A}\nodeD_{j}^{B}} - \sum_{i}w_{i}^{A}\sum_{j}w_{j}^{B}\E{\delta_{i}^{A}}\E{\nodeD_{i}^{A}}\E{\delta_{j}^{B}}\E{\nodeD_{j}^{B}} \\
                                      &= q^{2}\left(\sum_{i}w_{i}^{A}\sum_{j}w_{j}^{B}\E{\nodeD_{i}^{A}\nodeD_{j}^{B}} - \sum_{i}w_{i}^{A}\sum_{j}w_{j}^{B}\E{\nodeD_{i}^{A}}\E{\nodeD_{j}^{B}} \right) \\
                                      &= q^{2}\left(\sum_{i}w_{i}^{A}\sum_{j}w_{j}^{B}\left(\E{\nodeD_{i}^{A}\nodeD_{j}^{B}} - \E{\nodeD_{i}^{A}}\E{\nodeD_{j}^{B}} \right)\right) \\
                                      &= q^{2}\sum_{i}w_{i}^{A}\sum_{j}w_{j}^{B}\Cov{\nodeD_{i}^{A},\nodeD_{j}^{B}} \quad . \label{eq:cov-two-sum-nodes:final}
\end{align}

\paragraph{Product Nodes:}
As detailed in the paper's main body, we are unfortunately unable to provide a
closed-form solution of the covariance between two product nodes for an
arbitrary graph. This is due to the first expectation in the product node
covariance:
\begin{align}
  \label{eq:6}
  \Cov{\prodnode^{A},\prodnode^{B}} &= \Cov{\prod_{i}\nodeD_{i}^{A}, \prod_j \nodeD_{j}^{B}} \\
                                    &= \E{\prod_{i}\nodeD_{i}^{A} \prod_j \nodeD_{j}^{B}} - \E{\prod_{i}\nodeD_{i}^{A}} \E{\prod_j \nodeD_{j}^{B}}\\
                                    &= \E{\prod_{i}\nodeD_{i}^{A} \prod_j \nodeD_{j}^{B}} - \prod_{i}\E{\nodeD_{i}^{A}} \prod_j \E{\nodeD_{j}^{B}} \quad\label{eq:6:final} .
\end{align}
Further resolving the term $\E{\prod_{i}\nodeD_{i}^{A} \prod_j \nodeD_{j}^{B}}$
without additional knowledge about specific node substructures is not possible.
More specifically, pairs of $\nodeD_{i}^{A}$ and $\nodeD_{j}^{B}$ may not be
independent and may share a common subset of nodes, deeper down in the PC
structure. In this case, as alternatives, we have provided three possible
solutions to solve~\cref{eq:6} in the main body. For convenience, we briefly
re-iterate. The first solution is comprised of exploiting the circuit's
structural knowledge, if available. Alternatively, when such information is not
known, we can compute tractable bounds using the Cauchy-Schwarz inequality.
These two solutions are now presented in more detail in the following text.
Finally, recall that the third solution consists of ``augmenting'' the structure
of the circuit. Here, node copies for the shared nodes are used, as discussed in
paragraph \textbf{c)} of Section 3.2.3 of the main paper.

\subsubsection{Exploiting Structural Knowledge for Exact Covariance Computation}
\label{supp:structural_knowledge}
\paragraph{Tree-structured PCs:}
To simplify \cref{eq:6} we can directly exploit structural knowledge of the DAG.
A simple solution could be provided when product nodes $\prodnodeD^{A}$ and
$\prodnodeD^{B}$ are not common ancestors of any node, resulting in the
independence $\prodnodeD^{A} \ind \prodnodeD^{B}$ holding and thus
$\Covinline{\prodnodeD^{A}, \prodnodeD^{B}} = 0$. Such a constraint is always
given in tree-structured PCs. Notably, the most common structure learner for
SPNs such as LearnSPN~\citep{gens_2013}, ID-SPN~\citep{rooshenas_2014}, and
SVD-SPN \citep{adel_2015} pursue the simplicity bias and induce tree structures.

\paragraph{Random and Tensorized Binary Tree Structures:}
For random and tensorized (RAT) binary tree structures~\citep{peharz_2020}, we
need to account for the existence of cross-products, to compute the
variance of product nodes. For binary tree structures of this kind, we can thus
leverage the characteristic that product nodes factorize two completely
independent partitions. In the context of this particular structure with binary
partitioning, the variance for a RAT sum node, with its children having their
scope on graph partitions $L$ and $R$, can be rewritten in the following way:
\begin{align}
  \label{eq:2}
  \Var{\sumnode} &= \Var{\sum_{l}\sum_{r}w_{l,r}\delta_{l,r} \prodnodeD_{l,r}} \\
                 &= \sum_{l}\sum_{r} w_{l,r}^{2} \Var{\delta_{l,r}\prodnodeD_{l,r}} + \sum_{l,r}\sum_{\left(l',r'\right) \neq \left( l,r \right)} w_{l,r}w_{l',r'} \Cov{\prodnodeD_{l,r}, \prodnodeD_{l',r'}} \label{eq:2:final} \, .
\end{align}
Consequently, the covariance term between two product nodes in~\cref{eq:2:final}
can be simplified to:
\begin{align}
  \label{eq:5}
  \Cov{\prodnodeD_{l,r}, \prodnodeD_{l',r'}} &= \E{\prodnodeD_{l,r} \prodnodeD_{l',r'}} -  \E{\prodnodeD_{l,r}}\E{\prodnodeD_{l',r'}} \\
                                             &= \E{\sumnodeD^{L}_{l}\sumnodeD^{R}_{r} \sumnodeD^{L}_{l'} \sumnodeD^{R}_{r'}} - \E{\sumnodeD^{L}_{l}\sumnodeD^{R}_{r}}\E{\sumnodeD^{L}_{l'} \sumnodeD^{R}_{r'}} \, .
\end{align}
Since nodes $\sumnodeD^{L}$ and $\sumnodeD^{R}$ are from two different
partitions, they are independent, and the expectation terms can be separated:
\begin{alignat}{2}
  \label{eq:7}
  \Cov{\prodnodeD_{l,r}, \prodnodeD_{l',r'}} &= &&\E{\sumnode^{L}_{l} \sumnode^{L}_{l'}} \E{\sumnode^{R}_{r} \sumnode^{R}_{r'}} - \E{\sumnode^{L}_{l}}\E{\sumnode^{R}_{r}}\E{\sumnode^{R}_{r'}}\E{\sumnode^{L}_{l'}} \\
                                             &= &&\left( \E{\sumnode^{L}_{l}}\E{\sumnode^{L}_{l'}} + \Cov{\sumnode^{L}_{l}, \sumnode^{L}_{l'}} \right) \left( \E{\sumnode^{R}_{r}}\E{\sumnode^{R}_{r'}} + \Cov{\sumnode^{R}_{r}, \sumnode^{R}_{r'}} \right) \notag \\
                                             & &&-  \E{\sumnode^{L}_{l}}\E{\sumnode^{R}_{r}}\E{\sumnode^{L}_{l'}}\E{\sumnode^{R}_{r'}} \\
                                             &= &&\E{\sumnode^{L}_{l}}\E{\sumnode^{R}_{r}}\E{\sumnode^{L}_{l'}}\E{\sumnode^{R}_{r'}} \notag\\
                                             & &&+\Cov{\sumnode^{L}_{l}, \sumnode^{L}_{l'}} \E{\sumnode^{R}_{r}} \E{\sumnode^{R}_{r'}} \notag\\
                                             & &&+\Cov{\sumnode^{R}_{r}, \sumnode^{R}_{r'}} \E{\sumnode^{L}_{l}} \E{\sumnode^{L}_{l'}} \notag\\
                                             & &&+\Cov{\sumnode^{L}_{l}, \sumnode^{L}_{l'}} \Cov{\sumnode^{R}_{r}, \sumnode^{R}_{r'}} \notag\\
                                             & &&- \E{\sumnode^{L}_{l}}\E{\sumnode^{R}_{r}}\E{\sumnode^{L}_{l'}}\E{\sumnode^{R}_{r'}} \\
                                             &= &&\Cov{\sumnode^{L}_{l}, \sumnode^{L}_{l'}} \E{\sumnode^{R}_{r}} \E{\sumnode^{R}_{r'}} \notag\\
                                             & &&+\Cov{\sumnode^{R}_{r}, \sumnode^{R}_{r'}} \E{\sumnode^{L}_{l}} \E{\sumnode^{L}_{l'}} \notag\\
                                             & &&+\Cov{\sumnode^{L}_{l}, \sumnode^{L}_{l'}} \Cov{\sumnode^{R}_{r}, \sumnode^{R}_{r'}}  \, .
\end{alignat}
We can see that the covariance of two product nodes now depends on the
covariance of the input sum nodes of the same partition (i.e. $L$ or $R$), for
which we can plug in \cref{eq:cov-sum-node-children:last}.

\subsubsection{It's Somewhere in Here -- Covariance Bounds}
\label{covariance_bounds}
As previously described, knowledge about the specific PC structure can
facilitate the covariance computation that otherwise could result in a
combinatorial explosion. When such knowledge is not available, we can
alternatively obtain a lower and upper bound of the covariance in a tractable
way, making use of the Cauchy-Schwarz inequality:
\begin{alignat}{2}
  \label{supp:eq:cauchy-schwarz}
  &\Cov{\ndi, \ndj}^{2} &&\leq \Var{\ndi}\Var{\ndj} \\
  \Leftrightarrow \quad &\Cov{\ndi, \ndj} &&\in \left[ - \sqrt{\Var{\ndi}\Var{\ndj}}, + \sqrt{\Var{\ndi}\Var{\ndj}} \right] \quad . \label{supp:eq:cauchy-schwarz:last}
\end{alignat}

\subsubsection{Leaf Nodes}
\label{sec:tdi:leaf-nodes}
Finally, as the leaf nodes are free of any dropout Bernoulli variables, their
expectation, variance, and covariance degrade to the leaf node value and zero
respectively, i.e.:
\begin{align}
  \E{\leafnode} & = \leafnode \, , \\
  \Var{\leafnode} & = 0 \, , \\
  \Cov{\leafnode_{i}, \leafnode_{j}} & = 0 \, .
\end{align}

\subsubsection{Classification Uncertainty}
\label{sec:tdi:classification}
For classification in PCs, for a given sample $\x \sim \X$, it is common to
obtain the data conditional class confidence $p(y_{i} | \x)$ by using Bayes'
rule with the class conditional root nodes $p(\x | y_{i})$ and the class priors
$p(y_{i}) = \ci$. That is, every class $y_{i}$ has a corresponding root node
$\si$, representing $p(\x | y_{i})$. Following the earlier sections'
elaborations, we can then similarly derive the variance as an uncertainty proxy
in a classification context. Making use of Bayes' rule, we obtain the posterior
for class $y_{i}$ as follows:
\begin{align}
  \label{eq:posterior}
  p(y_{i} | \x) = \frac{p(\x | y_{i}) p(y_{i})}{\sum_{j} p(\x | y_{j}) p(y_{j})} = \frac{\si\ci}{\sum_{j}\sj\cj} \, .
\end{align}
Following existing notation, we will abbreviate the $i$-th root node with
$\si$ and the $i$-th class prior with $\ci$ for simplicity.

The expectation and variance of the posterior are that of a random variable
ratio, $\E{\frac{A}{B}}$ and $\Var{\frac{A}{B}}$, with $A = \si\ci$ and
$B = \sum_{j}\sj\cj$. This ratio is generally not well-defined, but can
be approximated with a second-order Taylor approximation~\citep{ratio_seltman}:
\begin{align}
  \E{\frac{A}{B}} &\approx \frac{\E{A}}{\E{B}} - \frac{\Cov{A, B}}{\left( \E{B} \right)^{2}} + \frac{\Var{B}\E{A}}{\left( \E{B} \right)^{3}} \, , \label{eq:exp-rv-ratio-suppl}\\
  \Var{\frac{A}{B}} &\approx \frac{\E{A}^{2}}{\E{B}^{2}} \left[ \frac{\Var{A}}{\E{A}^{2}} - 2 \frac{\Cov{A, B}}{\E{A}\E{B}} + \frac{\Var{B}}{\E{B}^{2}} \right] \, . \label{eq:var-rv-ratio-suppl}
\end{align}
We will now resolve every component of \cref{eq:exp-rv-ratio-suppl,eq:var-rv-ratio-suppl}.
The expectations can be expressed directly as:
\begin{align}
  \E{A} &= \E{\si\ci} = \E{\sumnode{i}}\ci \label{supp:eq:exp-x} \, , \\
  \E{B} &= \E{\sum_{j}\sj\cj} = \sum_{j}\Esj\cj \, .
          \label{supp:eq:exp-y}
\end{align}
For the variances we obtain:
\begin{align}
  \Var{A} &= \Var{\si\ci} = \Var{\si}\ci^{2} \label{supp:eq:var-x} \, , \\
  \Var{B} &= \Var{\sum_{j}\sj\cj} \\
          &= \sum_{j}\Var{\sj}\cj^{2} + \sum_{j_1 \neq j_2} \Cov{\sumnode_{j_{1}}, \sumnode_{j_{2}}}c_{j_{1}}c_{j_{2}} \, .
            \label{supp:eq:var-y}
\end{align}
Finally, following~\cref{eq:cov-two-sum-nodes}, the covariance term between a
root node and the sum of all root nodes can be decomposed as follows:
\begin{align}
  \label{supp:eq:cov-x-y}
  \Cov{A, B} &= \Cov{\si\ci, \sum_{j}\sj\cj} \\
             &= \E{\si \ci \cdot \sum_{j}\sj\cj} - \E{\si\ci}\E{\sum_{j}\sj\cj} \\
             &= \E{\sum_{j}\sj\si\ci\cj} - \E{\si\ci}\E{\sum_{j}\sj\cj} \\
             &= \ci\sum_{j} \cj \E{\si\sj} - \ci\Esi \sum_{j} \cj \Esj \\
             &= \ci\sum_{j} \cj \left( \Cov{\si,\sj} + \Esi\Esj \right) - \ci\Esi \sum_{j} \cj \Esj \\
             &= \ci \left(\left(\sum_{j} \cj \Cov{\si,\sj}\right) + \left(\Esi\sum_{j}\cj\Esj   \right) - \left(\Esi \sum_{j} \cj \Esj  \right)   \right) \\
             &= \ci \sum_{j} \cj \Cov{\si,\sj} \quad .
\end{align}
As previously described but repeated for emphasis, the last term
$\Cov{\si, \sj}$ can be resolved with one of the three methods
presented in Section 3.2.3 of the main paper, i.e., by exploiting structural
knowledge (see also~\cref{supp:structural_knowledge}), or by computing bounds
with \cref{supp:eq:cauchy-schwarz} of~\cref{covariance_bounds}, or by ``augmenting''
the structure with node copies of shared nodes.

While \cref{eq:var-rv-ratio-suppl} is seemingly simple, this particular
formulation implies statistical independence between $A$ and $B$. Since $B$ is
a sum over all $A$, this independence naturally does not hold. Therefore, the
solution given here is only an approximation of the true second-order Taylor
approximation. In the following we extend the formulation of
\cite{ratio_seltman} and take into account the dependencies between root nodes
$\si$ and their sum $\sum_{i}\si$.
For simplicity, we express \cref{eq:posterior} as a function $f$ of variables
$\sumnode_{1}, \dots, \sumnode_{C}$.
\begin{align}
  \fn{f}{\sumnode_{1}, \dots, \sumnode_{C} } = \fn{f}{ \mbtheta } + \sum_{i} \frac{\partial}{\partial \si} \fn{f}{ \mbtheta } \left( \si - \theta_{i} \right) + \sum_{i}\sum_{j} \frac{\partial^{2}}{\partial \si \partial \sj} \fn{f}{\mbtheta} \left(\si - \theta_{i} \right) \left( \sj - \theta_{j} \right) + R \quad ,
\end{align}
where $R$ is the remainder of smaller orders. To abbreviate equations, we introduce $Z = \sum_{j} \sj c_{j}$ and
$Z_{\backslash i} = \sum_{j \neq i} \sj c_{j}$.
To resolve the first-order term, we need the first partial derivative at $\si$. %
\newcommand{\Zni}[0]{Z_{\backslash i}}
\begin{align}
  \frac{\partial}{\partial \si} \fn{f}{\sumnode_{1}, \dots, \sumnode_{C}} &= \frac{Z \frac{\partial \si c_{i}}{\partial \si} - \si c_{i} \frac{\partial}{\partial \si} Z}{Z^{2}} \\
 &= \frac{ c_{i}Z  - \si c_{i}^{2} }{Z^{2}} \\
 &= c_{i} \frac{ Z  - \si c_{i} }{Z^{2}} \\
 &=  c_{i} \frac{ \Zni  }{ Z^{2}}  \label{eq:first-partial-derivative} \quad .
\end{align}
We continue with the second partial derivative at $S_{k}$ and make a distinction
for $k = i$ and $k \neq i$. For the second partial derivative at $S_{i}$ we get
\begin{align}
  \frac{\partial^{2}}{\partial \si \partial \si} \fn{f}{\sumnode_{1}, \dots, \sumnode_{C} } &= \frac{Z^{2} \left( \frac{\partial}{\partial \si} \Zni \right) - \Zni \left( \frac{\partial}{\partial \si} Z^{2} \right)}{Z^{4}} \\
  &= \frac{- 2 c_{i} \Zni Z}{Z^{4}} \\
  &= - 2 c_{i} \frac{\Zni}{Z^{3}} \quad . \label{eq:second-partial-derivative-i}
\end{align}
For the second partial derivative at $S_{k}$ with $k \neq i$ we get
\begin{align}
  \frac{\partial^{2}}{\partial \si \partial \sk} \fn{f}{\sumnode_{1}, \dots, \sumnode_{C} } &= \frac{Z^{2} \left( \frac{\partial}{\partial \sk} \Zni \right) - \Zni \left( \frac{\partial}{\partial \sk} Z^{2} \right)}{Z^{4}} \\
  &= \frac{c_{k} Z^{2} - 2 c_{k} Z \Zni}{Z^{4}} \\
  &= \frac{c_{k} Z - 2 c_{k} \Zni}{Z^{3}} \\
  &= c_{k} \frac{Z - 2 \Zni}{Z^{3}} \\
  &= c_{k} \frac{Z - 2 \left( Z - \si c_{i} \right)}{Z^{3}} \\
  &= c_{k} \frac{ - Z -  2 \si c_{i} }{Z ^{3}} \\
  &= - c_{k} \frac{ Z +  2 \si c_{i} }{Z^{3}} \label{eq:second-partial-derivative-k} \quad .
\end{align}
Using
\cref{eq:first-partial-derivative,eq:second-partial-derivative-i,eq:second-partial-derivative-k},
and $\theta_{i} = \Esi$ we obtain
\begin{multline}
  \fn{f}{\sumnode_{1}, \dots, \sumnode_{C} } = \fn{f}{\E{\sumnode_{1}}, \dots, \E{\sumnode_{C}}} + \sum_{i} \left( \frac{c_{i} \E{\Zni}}{ \E{Z}^{2} } \right) \left( \si - \Esi \right) \\
  + \sum_{i} \Biggl( \sum_{j \neq i} \left( - c_{j} \frac{\E{Z} + 2 \Esi c_{i}}{\E{Z}^{3}} \left( \si - \Esi \right) \left( \sj - \Esj \right) \right) \\
  - 2 c_{i} \frac{\E{\Zni}}{\E{Z}^{3}}  \left( \si - \Esi \right)^{2}\Biggr) + R \quad .
\end{multline}
Now that we have derived the second order Taylor approximation of
\cref{eq:posterior}, we need to take the expectation and variance of $f$. We
begin with the expectation as follows
\begin{multline}
  \E{\fn{f}{\sumnode_{1}, \dots, \sumnode_{C} }} = \fn{f}{\E{\sumnode_{1}}, \dots, \E{\sumnode_{C}}} + \sum_{i} \left( \frac{c_{i} \E{Z}}{\E{Z}^{2}} \right) \overbrace{\E{\si - \Esi}}^{= \Ens{\si} - \Ens{\si} = 0} \\
  + \sum_{i} \Biggl( \sum_{j \neq i} \left( - c_{j} \frac{\E{Z} + 2\Esi c_{i}}{ \E{Z}^{3} } \overbrace{\E{ \left( \si - \Esi \right) \left( \sj - \Esj \right)  }}^{= \Covns{\si,\sj}}\right)  \\
  - 2 c_{i} \frac{\E{\Zni}}{ \E{Z}^{3}} \overbrace{\E{\left(\si - \Esi\right)^{2}}}^{= \Varns{\si}} \Biggr) \quad ,
\end{multline}
which simplifies to
\begin{equation}
  \label{eq:second-order-taylor-exp-final}
  \E{\fn{f}{\sumnode_{1}, \dots, \sumnode_{C} }} = \fn{f}{\E{\sumnode_{1}}, \dots, \E{\sumnode_{C}}} + \sum_{i} \left( \sum_{j \neq i} \left( - c_{j} \frac{\E{Z} + 2\Esi c_{i}}{ \E{Z}^{3} } \Cov{\si,\sj}\right)
  - 2 c_{i} \frac{\E{\Zni}}{  \E{Z}^{3}} \Var{\si} \right) \, .
\end{equation}
We continue with the variance of $f$
\begin{multline}
  \Var{\fn{f}{\sumnode_{1}, \dots, \sumnode_{C} }} = \Var{\fn{f}{\E{\sumnode_{1}}, \dots, \E{\sumnode_{C}}}} + \sum_{i} \Var{c_{i} \frac{\E{\Zni}}{\E{Z}^{2}} \left( \si - \E{\si} \right) } \\
  + \sum_{i} \Biggl( \sum_{j \neq i} \left( - c_{j} \frac{2\E{\si}c_{i} + \E{Z}}{\E{Z}^{3}} \right)^{2} \Var{\left( \si - \E{\si} \right) \left( \sj - \E{\sj} \right)} \\
  + \left( - 2 c_{i} \frac{\E{\Zni}}{\E{Z}^{3}} \right) \Var{\left( \si - \E{\si} \right)^{2}} \Biggr) \quad .
\end{multline}
Resolving each variance term simplifies the equation to
\begin{multline}
  \label{eq:second-order-taylor-var-final}
  \Var{\fn{f}{\sumnode_{1}, \dots, \sumnode_{C} }} = \sum_{i} \left( c_{i} \frac{\E{\Zni}}{\E{Z}^{2}} \right)^{2} \Var{\si} \\
  + \sum_{i} \Biggl( \sum_{j \neq i} \left( - c_{j} \frac{2\E{\si}c_{i} + \E{Z}}{\E{Z}^{3}} \right)^{2} \left(\Var{\si\sj} - \E{\si}^{2}\Var{\sj} - \E{\sj}^{2}\Var{\si}\right) \\
  + \left( - 2 c_{i} \frac{\E{\Zni}}{\E{Z}^{3}} \right) \left(\Var{\si^{2}} - 4\E{\si}^{2}\Var{\si}\right) \Biggr) \quad .
\end{multline}
We now end up with a variance term that needs further resolution: the product of
two sum nodes.
\newcommand{\wi}[1]{w_{#1}^{\sumnode_i}}
\newcommand{\wj}[1]{w_{#1}^{\sumnode_j}}
\newcommand{\Ni}[1]{\node_{#1}^{\sumnode_i}}
\newcommand{\Nj}[1]{\node_{#1}^{\sumnode_j}}
\begin{align}
  \label{eq:var-sum-sum}
  \Var{\si\sj} &= \Var{\sum_{k_{i}}\wi{k_{i}}\Ni{k_{i}}  \sum_{k_{j}}\wi{k_{i}}\Ni{k_{i}}} \\
  &= \Var{\wi{1}\wj{1}\Ni{1}\Nj{1} + \dots + \wi{1}\wj{\left|\sj\right|}\Ni{1}\Nj{\left|\sj\right|} \dots + \wi{\left|\si\right|}\wj{\left|\sj\right|}\Ni{\left|\si\right|}\Nj{\left|\sj\right|}} \\
  &= \sum_{k_{i}}\sum_{k_{j}} \Var{\wi{k_{i}}\wj{k_{j}}\Ni{k_{i}}\Nj{k_{j}}} \\
  &= \sum_{k_{i}}\sum_{k_{j}} \left(\wi{k_{i}}\wj{k_{j}}\right)^{2} \Var{\Ni{k_{i}}\Nj{k_{j}}} \quad .
\end{align}
Analogous to the discussions in \cref{sec:tdi:covariance}, the complexity of computing the variance of the product of two product nodes can be addressed through the incorporation of supplementary structural knowledge (refer to \cref{supp:structural_knowledge} for further details) or by utilizing crude approximations that induce local independence. An example of such an approximation is expressed as $\Var{\Ni{k_{i}}\Nj{k_{j}}} = \Var{\Ni{k_{i}}}\Var{\Nj{k_{j}}}$. With \cref{eq:second-order-taylor-exp-final,eq:second-order-taylor-var-final,eq:var-sum-sum} we have now reduced the second-order Taylor posterior approximation to quantities that we are able to compute from a single bottom up pass through the PC graph.

\newpage

\section{Tractable Dropout Inference Pseudocode Algorithm}
Analogously to the conventional probabilistic inference on PCs, TDI is performed
by evaluating the circuit bottom-up. For clarity and completeness, we present
the pseudocode of the bottom-up pass with TDI in
\cref{algo:tractable_dropout_inference}. While the bottom-up pass can be
implemented in a variety of different ways, we have decided on a recursive
version in the presented pseudocode for shortness and clarity. We start by
initializing empty maps \texttt{E}, \texttt{Var}, and \texttt{Cov} that map nodes to
their expectation, variance, and covariance values. Given a node $\nodeD$, some
data sample $\mbx$, and the dropout probability $p$, we then now traverse the
graph below $\nodeD$ and require, that the \texttt{tdi} procedure has been
called on all children, ensuring that the \texttt{E}, \texttt{Var},  and
\texttt{Cov} of all $\nodeD_{i} \in \ch\left(\nodeD\right)$ are populated. We
then continue to compute the covariance between all child nodes of $\nodeD$,
followed by the expectation and finally the variance of $\nodeD$. All three
procedure calls, \texttt{expectation}, \texttt{variance}, and
\texttt{covariance}, refer to the equations given in the main body in
\cref{sec:tdi:derivation} for sum, product, and leaf nodes respectively.

\begin{algorithm}[h]
    \caption{Tractable Dropout Inference: \texttt{tdi}$\left(\nodeD, \mbx, p\right)$}
    \label{algo:tractable_dropout_inference}
    \begin{algorithmic} %
      \Require PC root node $\nodeD$; Data sample $\mbx$; Dropout probability
      $p$; Empty maps \texttt{E}, \texttt{Var}, \texttt{Cov}.
      \ForAll{$\ndi \in \ch\left(\nodeD\right)$}
        \State{\texttt{tdi}$\left(\nodeD_{i}, \mbx, p\right)$}
        \Comment{Populate maps for child nodes recursively}
      \EndFor

      \ForAll{$\ndi \in \ch\left(\nodeD\right)$}
        \ForAll{$\ndj \in \ch\left(\nodeD\right)$}
          \State{$\text{\texttt{Cov}}\!\left[ \ndi, \ndj \right] = \text{\texttt{covariance}}\left(\nodeD_{i}, \nodeD_{j}, \mbx, p\right)$}
          \Comment{See \cref{eq:sum-covariance,eq:prod-covariance,eq:leaf-exp-var-cov}}
        \EndFor
      \EndFor

      \State{$\text{\texttt{E}}\!\left[ \nodeD \right] = \text{\texttt{expectation}}\left( \nodeD, \mbx, p \right)$}
      \Comment{See \cref{eq:sum-expectation,eq:prod-expectation,eq:leaf-exp-var-cov}}
      \State{$\text{\texttt{Var}}\!\left[ \nodeD \right] = \text{\texttt{variance}}\left( \nodeD, \mbx, p \right)$}
      \Comment{See \cref{eq:sum-variance,eq:prod-variance,eq:leaf-exp-var-cov}}
    \end{algorithmic}
  \end{algorithm}

  \section{Experimental Setup}
  \label{supp:appendix_b}
  For our experiments, we implemented TDI based on the publicly-available
  PyTorch implementation of
  RAT-SPNs\footnote{\url{https://github.com/SPFlow/SPFlow/tree/master/src/spn/experiments/RandomSPNs_layerwise}}
  in the \emph{SPFlow} software package ~\citep{molina_2019}.
  Our code is available at \url{https://github.com/ml-research/tractable-dropout-inference}.
  All our models were trained for 200 epochs with the Adam optimizer
  \citep{kingma2015adam}, employing a batch size of 200 and a learning rate of
  1e-3. To assess the robustness of probabilistic circuits as probabilistic
  discriminators, we have run all our experiments with the RAT hyperparameter
  $\lambda$ set to $1$. The latter corresponds to maximizing the class
  conditional likelihood of a sample computed by the head attributed to the
  observed label. Regarding the RAT-SPN graph, we make use of the following
  hyperparameters: $S=20, I=20, D=5, R=5$ with Gaussian leaves. This
  configuration returns circuits with 1.65M edges and 1.83M learnable parameters
  for SVHN (614k are Gaussian leaf parameters), and 1.42M edges and 1.38M
  parameters for MNIST (157k are Gaussian leaf parameters).
  We emphasize that all previously described hyperparameter choices correspond
  to standard, previously considered, choices in the literature. No additional
  or excessive hyperparameter tuning has been conducted on our side. We refer to
  the original work of \cite{peharz_2020} for further details on RAT-SPN
  training and its respective hyperparameters.

  For our TDI experiments, we select the dropout parameter $p$ to be $0.1$ for
  SVHN and $0.2$ for MNIST. This range is adequate to preserve accuracy and
  simultaneously provide valuable uncertainty estimates. Note how such a choice
  is consistent and not fundamentally different from dropout probabilities in
  neural networks (i.e. typically on the same scale in every layer, unless it is
  the last layer, where sometimes a $p$ of up to 0.5 can be found in the
  literature). However, we further note that probabilistic circuits are
  typically sparser than their neural counterparts. A small dropout parameter is
  thus generally sufficient but nevertheless remains a hyperparameter to be
  chosen.

\section{PCs with TDI can detect OOD data}
In the main paper, we have shown that PCs tend to be overconfident on OOD
instances while PCs + TDI can overcome this challenge and can detect
when an instance comes from a completely different distribution compared to what
observed during training. This facilitates the separation between ID and OOD
instances with a threshold, a peculiarity that is particularly useful also in
practice in many domains. Here, we provide a complementary view of this ability
of PCs with TDI. We show the predictive entropy in
\cref{fig:predictive-entropy-uncertainty:entropy} and the predictive uncertainty
in \cref{fig:predictive-entropy-uncertainty:uncertainty} obtained on the ID data
and on several OOD datasets.

\begin{figure}[h]
  \centering
  \begin{subfigure}[t]{0.5\linewidth}
    \centering
    \includegraphics[width=1.0\linewidth]{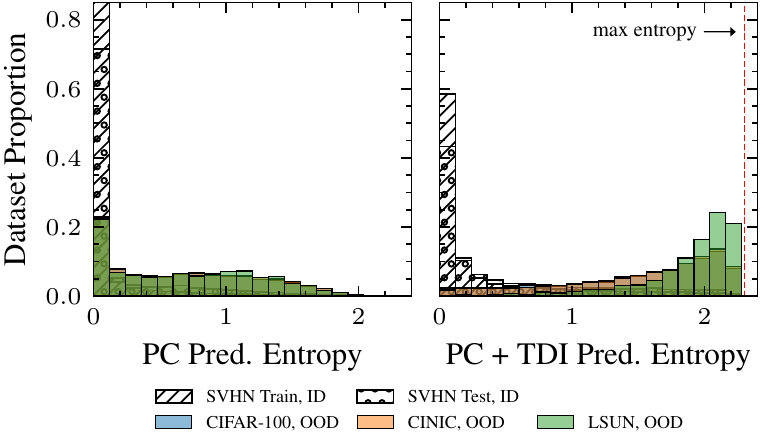}
    \caption{}
    \label{fig:predictive-entropy-uncertainty:entropy}
  \end{subfigure}%
  \begin{subfigure}[t]{0.5\linewidth}
    \centering \includegraphics[width=1.0\linewidth]{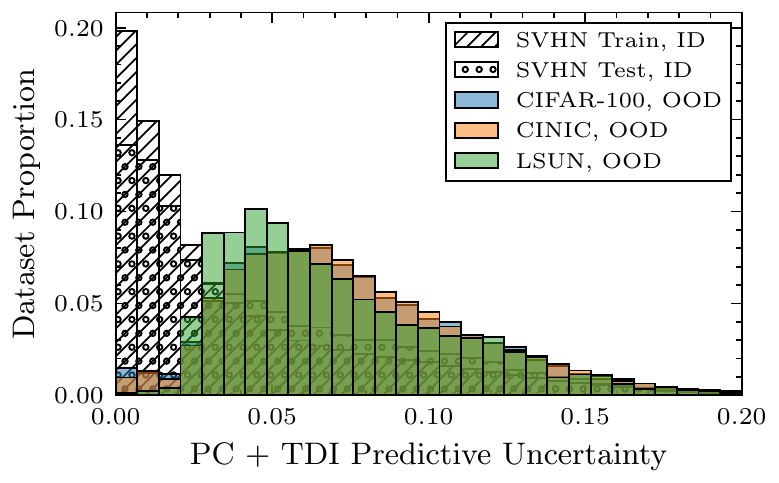}
    \caption{}
    \label{fig:predictive-entropy-uncertainty:uncertainty}
  \end{subfigure}%
  \caption{
  (\subref{fig:predictive-entropy-uncertainty:entropy}) Histograms for the predictive entropy of PCs and PCs + TDI on ID (shaded) and OOD (colored) datasets in a classification context.
  PCs (left panel) assign similar, largely indistinguishable predictive entropy to OOD and ID data. In contrast, PCs + TDI (right panel) provides high uncertainty on OOD data, pushing the predictive entropy close to the obtainable maximum and effectively separating them from the known ID dataset.
  (\subref{fig:predictive-entropy-uncertainty:uncertainty}) Predictive
  uncertainty (standard deviation as square root of \cref{eq:var-rv-ratio}) for the predictions
  provided by PC + TDI on ID (shaded) and OOD data (colored). Complementing the
  illustration of predictive entropy in
  (\subref{fig:predictive-entropy-uncertainty:entropy}), the values support the
  picture that PCs + TDI are more unsure about OOD data and can thus make a
  distinction. }
\label{fig:predictive-entropy-uncertainty}
\end{figure}

\section{PCs with TDI Are More Robust To Corruptions}
\label{supp:appendix_c}
We have applied 15 non-trivial natural and synthetic corruptions with respective
five levels of severity, as shown in the main body's experimental evidence
section. For convenience, we re-iterate that the latter are commonly employed in
the literature~\citep{hendrycks_2019} to test models' robustness. More
specifically, corruptions have been used to demonstrate standard neural
networks' inability to effectively handle corrupted data. Since we have observed
an initial similar inability in PCs, they thus form a sensible test to evaluate
the robustness of PCs versus PCs + TDI. To provide some visual intuition, a
respective SVHN illustration is shown in~\cref{supp:fig:svhn_corruptions}.

In our main body, we have shown the performance of PCs and PCs + TDI for 4 such
corruptions and thus have provided empirical evidence for the robustness of PCs
with TDI and their ability to attribute increased uncertainty to corrupted data.
Here, we provide the full set of experiments for all 15 corruptions. Recall that
for a model to successfully detect the corruption, it should assign a
progressive increase in predictive entropy in accordance with the corruption
severity.

The full quantitative results of \cref{fig:svhn_corruptions_pred_entropy}
empirically affirm our conclusions drawn in the main body: PCs with TDI are
generally equally or more robust at various severity levels for all corruptions,
generally preserving equal or even higher predictive accuracy in several cases.
At the same time, the assigned predictive entropy is substantially larger with
an increasing level of corruption with TDI. Finally, we emphasize that in the
three cases where PCs with TDI do not provide a large and successively
increasing measure of entropy, i.e. zoom blur, pixelation, and jpeg compression,
their respective accuracy in \cref{fig:svhn_corruptions_pred_entropy} is almost
fully preserved. In other words, the model does not provide a large uncertainty
because it already provides a correct and robust prediction.

\begin{figure}[!h]
  \centering
  \includegraphics[width=0.6\linewidth]{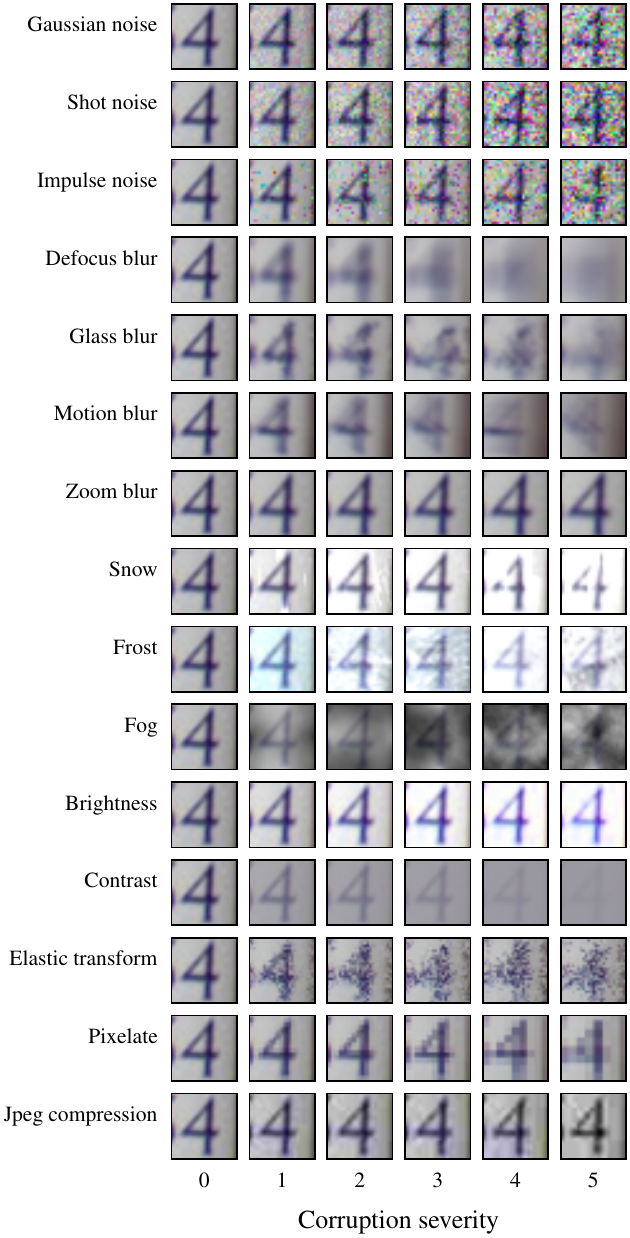}
  \caption{Visual illustration of the 15 different corruptions with five
    increasing levels of severity as introduced in~\cite{hendrycks_2019} on an
    SVHN test sample.}
  \label{supp:fig:svhn_corruptions}
\end{figure}

\begin{figure*}[!h]
  \centering \includegraphics[width=1.0\linewidth]{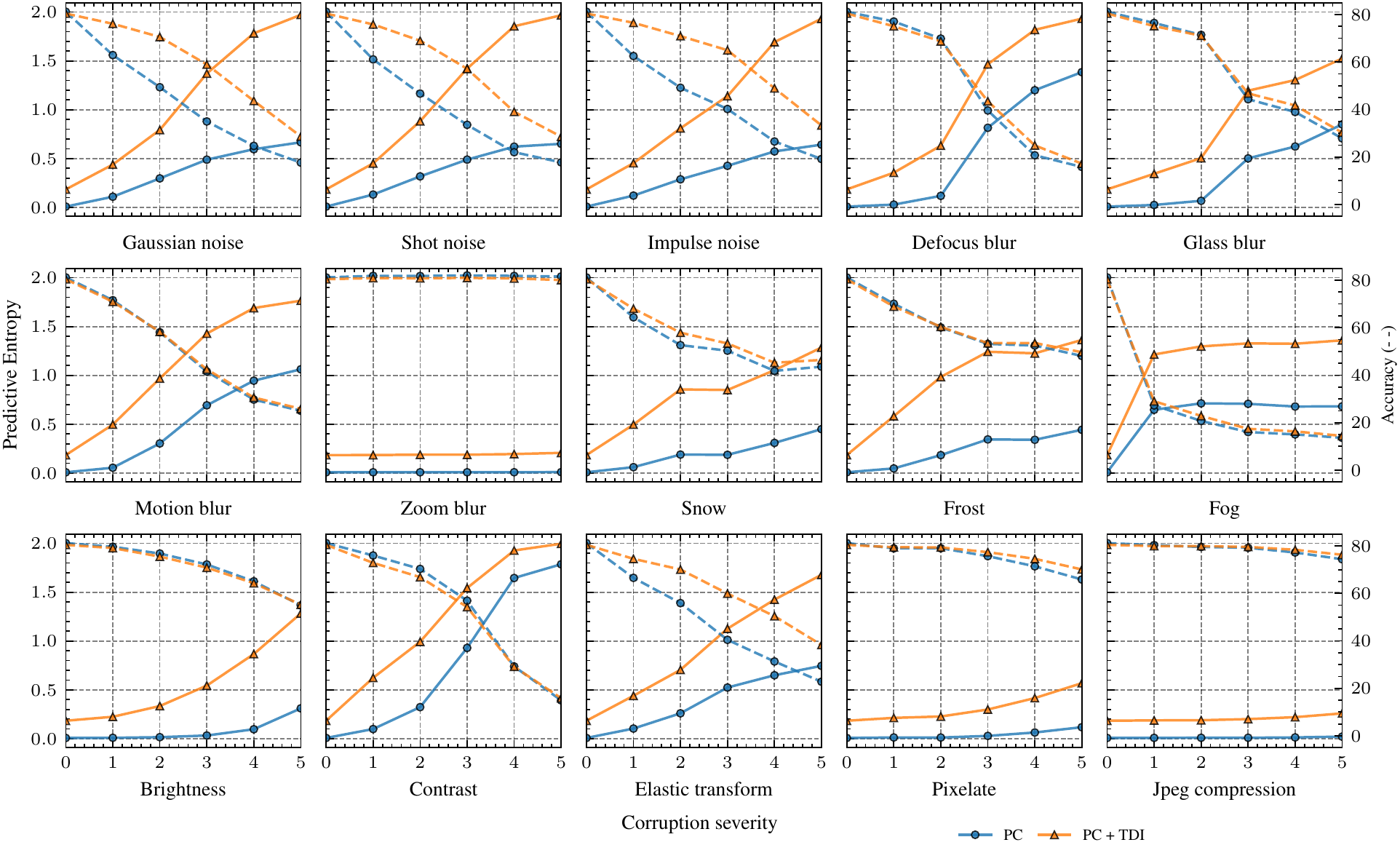}
  \caption{ Predictive entropy (left y-axis) and accuracy (right y-axis) of PCs
    (blue curve, circle markers) and PCs + TDI (orange, triangles) for
    increasingly corrupted SVHN data. 15 natural and synthetic corruptions are
    introduced at five severity levels. PCs with TDI can detect the distribution
    shift by assigning higher predictive entropy with increasing severity, while
    at the same time being more robust in predictive accuracy against the
    corruption, compared to PCs. We emphasize that in the three cases where PCs
    + TDI do not provide a large and successively increasing measure of entropy,
    i.e. zoom blur, pixelation, and jpeg compression, their respective accuracy
    (dashed) is almost fully preserved. The model is already correct and
    certain.}
  \label{fig:svhn_corruptions_pred_entropy}
\end{figure*}

\section{Societal Impact}
We believe that quantification of model uncertainty, as a direction to
contribute to overall robustness, can have a broad positive societal impact.
Colloquially speaking, an indication of when to ``trust'' the model is a
valuable tool for users and practitioners, especially in safety-critical
applications. That noted, gauging model uncertainty does not absolve users from
careful considerations of other potentially harmful effects, particularly, the
involvement of various forms of bias through training data, as these are then
considered ``certain'' by definition.
\fi

\end{document}